\documentclass[11pt]{article}

\usepackage[final]{acl}

\usepackage{times}
\usepackage{latexsym}

\usepackage[T1]{fontenc}

\usepackage[utf8]{inputenc}

\usepackage{microtype}

\usepackage{inconsolata}

\usepackage{graphicx}

\usepackage{amsmath}
\usepackage{hyperref}
\usepackage{url}
\usepackage{booktabs}
\usepackage{graphicx} 
\usepackage[table,x11names]{xcolor}
\usepackage{multirow}
\usepackage{pifont}
\usepackage{wrapfig} 
\usepackage{amssymb}
\usepackage{amsthm}
\usepackage{caption}
\usepackage{subcaption}
\usepackage{colortbl}
\usepackage{float}

\usepackage{fancyhdr}

\definecolor{mycolor}{rgb}{0, 0, 0} 
\newcommand{\coloredtext}[1]{\textcolor{mycolor}{#1}}

\usepackage{xspace}
\DeclareRobustCommand{\cf}{cf.\@\xspace}

\usepackage{algorithm}
\usepackage{algpseudocode}

\usepackage{xcolor}

\usepackage[most]{tcolorbox}
\tcbuselibrary{listings, skins} 

\tcbset{
  promptstyle/.style={
    enhanced,
    breakable,
    colframe=boxheader,
    colback=boxbody,
    coltitle=white,
    fonttitle=\bfseries\large,
    sharp corners,
    boxrule=0.6pt,
    left=6pt,right=6pt,top=6pt,bottom=6pt,
  }
}

\lstdefinestyle{recreatepython}{
    language=Python,
    basicstyle=\ttfamily\scriptsize, 
    keywordstyle=\color{blue!60!black}\bfseries,
    stringstyle=\color{green!40!black},
    commentstyle=\color{gray}\itshape,
    morekeywords={self, Path, replace_method}, 
    showstringspaces=false,
    breaklines=true,      
    breakatwhitespace=true,
    tabsize=4,
    frame=none,           
    aboveskip=0pt,
    belowskip=0pt
}

\lstdefinestyle{yamlstyle}{
    basicstyle=\ttfamily\scriptsize,
    keywords={content, created, id, tags, title}, 
    keywordstyle=\color{blue!60!black}\bfseries,
    stringstyle=\color{green!40!black},
    commentstyle=\color{gray}\itshape,
    morecomment=[l]{\#},
    breaklines=true,
    breakindent=1em,
    frame=none,
    aboveskip=0pt,
    belowskip=0pt,
    showstringspaces=false
}

\makeatletter
\newcommand{\appendixtableofcontents}{%
  \section*{Contents}
    \begingroup
    \small
    \setlength{\parskip}{2ex plus 0.2ex minus 0.2ex}%
    \setlength{\itemsep}{0.4ex}%
    \@starttoc{app}%
  \endgroup
}

\newcommand{\appsection}[1]{%
  \section{#1}%
  \addcontentsline{app}{section}{\protect\numberline{\thesection}#1}%
}

\newcommand{\appsubsection}[1]{%
  \subsection{#1}%
  \addcontentsline{app}{subsection}{\protect\numberline{\thesubsection}#1}%
}
\makeatother

\definecolor{boxheader}{RGB}{0, 128, 128} 
\definecolor{boxbody}{RGB}{240, 248, 255} 

\newtcolorbox{promptbox}[1][]{
  colframe=boxheader,      
  colback=boxbody,         
  coltitle=white,          
  fonttitle=\bfseries\large, 
  title={#1},              
  fontupper=\small\ttfamily, 
  boxrule=0.8mm,           
  arc=3mm,                 
  left=5pt, right=5pt, top=5pt, bottom=5pt, 
  breakable,
  enhanced,
}

\usepackage{enumitem}
\usepackage{listings}

\newtcolorbox{codepromptbox}[1]{
  colback=cyan!5!white,      
  colframe=teal!80!black,    
  title=#1,                  
  fonttitle=\bfseries,       
  left=3pt, right=3pt, top=3pt, bottom=3pt, 
  boxrule=1pt,               
  arc=2mm,                   
}

\title{ReCreate: Reasoning and Creating Domain Agents Driven by Experience}

\author{\textbf{Zhezheng Hao}$^{1}$\footnotemark[3]\quad
  \textbf{Hong Wang}$^{2}$\quad
  \textbf{Jian Luo}$^{3}$\quad
  \textbf{Jianqing Zhang}$^{4}$\quad
  \textbf{Yuyan Zhou}$^{2}$\quad
  \\[3pt]
  \textbf{Qiang Lin}$^{2}$\quad
  \textbf{Can Wang}$^{1,5}$\footnotemark[3]\quad
  \textbf{Hande Dong}$^{2}$\setcounter{footnote}{1}\thanks{Corresponding Authors}\quad
  \textbf{Jiawei Chen}$^{1,5}$\footnotemark[2]\thanks{State Key Laboratory of Blockchain and Data Security, Zhejiang University}\\[3pt]
  $^{1}$ Zhejiang University \quad
  $^{2}$ Workbuddy Team, Tencent \quad  \\[3pt]
  $^{3}$ Independent Researcher \quad
  $^{4}$ Shanghai Jiao Tong University \\[3pt]
  $^{5}$ Hangzhou High-Tech Zone (Binjiang) Institute of Blockchain and Data Security \\[3pt]
    \small Emails: \url{haozhezheng@outlook.com}, 
  \small\url{donghd66@gmail.com}, 
  \small\url{sleepyhunt@zju.edu.cn}
}

\begin{document}

\maketitle

\begin{abstract}
\vspace{-1mm}
Large Language Model (LLM) agents are reshaping the industrial landscape. However, most practical agents still rely on human-designed scaffolds, making their creation and adaptation across diverse domains labor-intensive. This situation poses a central question: can we automatically create and evolve domain agents through interaction experience in real-world environments? While recent approaches have sought to automate agent creation, several treat scaffold optimization as a black-box procedure guided primarily by final performance metrics. Such strategies overlook critical evidence explaining why an agent succeeds or fails and often incur high computational costs. To address these limitations, we propose ReCreate, an experience-driven framework for creating and evolving specialized domain agents. ReCreate systematically leverages agent interaction histories to identify the causes of success and failure and guide targeted scaffold updates. Specifically, we introduce an agent-as-optimizer paradigm in which an optimizer agent actively inspects execution experience and iteratively refines the task agent’s scaffold through three key components: (i) an experience storage and retrieval mechanism for on-demand evidence inspection; (ii) a reasoning–creating synergy pipeline that translates execution evidence into updates to prompts, workflows, tools, and memory; and (iii) hierarchical updates that consolidate instance-level refinements into reusable domain-level improvements. Experiments across thirteen evaluation settings spanning four domains show that ReCreate consistently outperforms human-designed agents and existing automated agent generation and self-evolving baselines, even when evolving from minimal seed scaffolds.

\end{abstract}

\section{Introduction}

As the capabilities of large language models (LLMs) continue to improve~\citep{gpt52, gemini3, anthropic2025claude45, liu2025deepseek, chang2024survey}, LLM-based agents have demonstrated striking competence on complex, long-horizon tasks, such as software engineering~\citep{yang2025code, yang2024swe}, scientific discovery~\citep{tang2025ai, weng2024cycleresearcher}, GUI operation~\citep{zhang2026dontactblindlyrobust, li2026whatsmissingscreentoactionuiintheloop} and web navigation~\citep{he2024webvoyager, team2025tongyi}.
These LLM-based agent systems are typically built on \emph{agent scaffolds} that specify how the model is prompted, how tasks are decomposed and executed, and how tools and environment feedback are integrated~\citep{luo2025large, xi2025rise}.
The success of these LLM-based agents shows that designing agent scaffolds is a critical step toward unlocking raw LLMs' capabilities and grounding raw LLMs in practical deployments~\citep{wang2024empowering, li2025deepagent}.

\begin{figure}[t]
    \centering
    \includegraphics[width=0.95\linewidth]{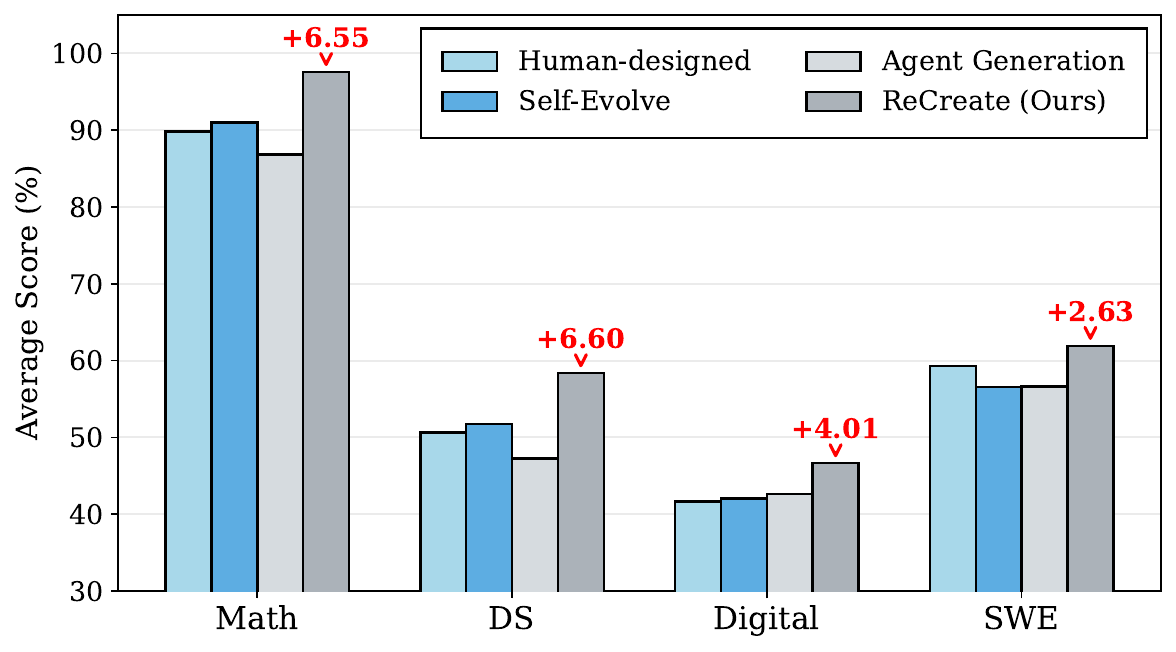}
    \vspace{-0.5em}
    \caption{Overall performance across four domains with \texttt{gpt-5-mini}.
    Bars report domain-level average scores of three method families and the proposed ReCreate.
    Red annotations indicate ReCreate's improvement over the runner-up in each domain.}
    \label{The overview of ReCreate.}
    \vspace{-2em}
\end{figure}

Yet in practice, LLM agents still rely on human-designed scaffolds, since different domains call for distinct knowledge and priors~\citep{xia2024agentless, ma2024sciagent, li2025codepde, li2025search}.
Manually designing and refining scaffolds is labor-intensive and hard to scale to numerous open-world scenarios (cf. Appendix~\ref{sec:human_scaffold_limits}).
This tension raises a central question: \textbf{can we automatically create specialized agents from minimal seeds and evolve them through interaction experience in real-world environments?}
In this work, we refer to this problem as \emph{domain agent creation} and approach it through experience-driven scaffold evolution.

A growing line of work studies automated agent generation, which replaces human-crafted scaffolds with a meta-agent that iteratively proposes, evaluates, and refines task-agent scaffolds~\citep{hu2024automated, shang2024agentsquare, zhang2024aflow, li2025agentswift, wang2025scoreflow}.
Scaffold optimization in these methods is typically driven by \textbf{performance metrics} (such as pass rates or LLM-judged scores), which are used to select and update candidate agents.
While this strategy has yielded promising progress, relying solely on performance metrics presents two limitations:
(1) Performance metrics do not provide evidence about \emph{why} and \emph{how} the agent succeeds or fails.
Consequently, agent optimization is typically treated as a black-box process, relying on exhaustive trial-and-error to uncover effective directions for scaffold improvement, thereby undermining both efficiency and effectiveness.
(2) Obtaining this metric value is often costly.
Each candidate scaffold typically requires large-scale evaluation to yield a stable and reliable performance measure.
For example, ADAS~\citep{hu2024automated} spends about $\$500$ for a single agent generation on ARC dataset~\citep{chollet2019measure} with only 20 task samples.

These limitations stem from treating agent creation and refinement as black-box optimization driven purely by performance scores.
Motivated by this, we shift towards a white-box optimization paradigm that leverages the agent's \textbf{interaction experience} --- including execution trajectories, evaluation logs, and environment state --- as primary evidence for scaffold refinement.
Such experience provides insight into \emph{why} an agent succeeds or fails, offering semantic and concrete evidence for adding rules, updating tools, and revising workflows (Section~\ref{sec:motivation} provides illustrative examples).

To implement this idea, we introduce \textit{ReCreate}, an experience-driven framework for automatically creating and evolving specialized domain agents.
ReCreate explicitly exploits rich interaction experience to guide iterative scaffold updates, with updated scaffolds generating new experience for subsequent refinement.
However, this process faces three key challenges:
(i) The large scale of interaction and environment information is challenging for LLMs to tackle;
(ii) Extracting meaningful evidence from complex experiences and converting it into suitable scaffold updates is inherently nontrivial;
(iii) Scaffold updates may easily overfit to single-task experiences rather than capture broader domain patterns, thus hindering domain-level generalization.
We address these challenges through an \textit{agent-as-optimizer} design comprising three components:
(1) an experience storage and retrieval mechanism that enables on-demand evidence inspection within the ReCreate environment;
(2) a reasoning--creating synergy pipeline that maps execution evidence into scaffold updates;
(3) a hierarchical update mechanism that aggregates instance-level refinements into reusable domain-level patterns.
Empirical evaluations across diverse domains confirm that ReCreate evolves minimal seed scaffolds into powerful specialized agents, achieving strong domain performance at a lower optimization cost than automated agent generation baselines.

Overall, the major contributions in this work are:
\setlength{\leftmargini}{0.5em}
\begin{itemize}

    \item \emph{The ReCreate framework:}
    We highlight the importance of interaction experience and propose ReCreate, a framework that automatically creates and iteratively refines agent scaffolds by learning from execution experience rather than relying solely on performance metrics.

    \item \emph{Agent-as-optimizer design:}
    Within ReCreate, we introduce an agent-as-optimizer design that efficiently inspects large-scale experience logs, grounds iterative scaffold updates in execution evidence, and extracts reusable domain-level patterns.

    \item \emph{Comprehensive evaluation:}
    We evaluate ReCreate across thirteen evaluation settings spanning four domains, showing consistent performance gains over human-designed agents and existing automated agent generation and self-evolving baselines.

\end{itemize}

\section{Preliminaries}
\subsection{Agent Scaffolds}
An LLM agent can be viewed as the composition of a base model $\phi$ and an agent scaffold $\mathcal{A}$~\citep{suzgun2024meta, xi2025rise}.
Formally, given a base LLM $\phi$, an agent scaffold $\mathcal{A}$ denotes the surrounding software layer that makes the base LLM $\phi$ executable in an environment~\citep{Raising, meirelesinfluence}.
To make agent scaffold editable, we systematically examined recent open-source, general-purpose agent scaffolds and abstracted their common design patterns~\citep{yang2024swe, wang2024openhands, openmanus2025}.
Based on their functions, we decompose $\mathcal{A}$ into the following complementary modular components:
\begin{itemize}
    \item \textbf{Role \& Object}:
    the system instruction that defines the agent’s identity, domain priors, and high-level goals;

    \item \textbf{Process \& Strategy}:
    the procedure that guides step-by-step reasoning, intermediate checks, and termination criteria;
    
    \item \textbf{Action \& Tool}:
    the action space exposed to the agent, implemented as reusable tools and scripts, including memory tools, search tools, etc;

    \item \textbf{Memory \& Retrieval}:
    the mechanism that controls how the agent stores, retrieves memory.
\end{itemize}
We therefore decompose agent scaffold $\mathcal{A}$ into a tuple
$\mathcal{A}=\bigl(
\mathcal{A}^{\text{role}},
\mathcal{A}^{\text{proc}},
\mathcal{A}^{\text{tool}},
\mathcal{A}^{\text{mem}}
\bigr)$, with each component editable.
In our implementation, other components, such as the action–observation format and error format, are ignored as they are non-essential to practical performance.

\subsection{Domain Agent Creation Problem}
\label{subsec:problem_def}
Many real-world domains lack expert-crafted agents, offering no ready-to-use workflows, rules, or tools.
In this scenario, available resources are limited to a base LLM $\phi$, a distribution $\mathcal{D}$ of verifiable tasks, and minimal domain information $\mathcal{I}$ (e.g., interfaces and constraints).
Formally, the domain agent creation problem seeks to construct a scaffold $\mathcal{A}$ from the tuple $(\phi,\mathcal{D},\mathcal{I})$ that transforms $\phi$ into a reliable agent for $\mathcal{D}$.
Unlike prompt tuning or tool learning, the goal is not adaptation to specific queries, but the creation of a plug-in agent that captures domain-level knowledge and generalizes to unseen tasks.









\section{Motivation: Experience Matters}\label{sec:motivation}
Our core motivation is that interaction experience, \coloredtext{which contains the full trajectory, execution results and evaluation results,} carries the agent’s action path and reasoning process.
This information can be leveraged to design stronger agent scaffolds.
\coloredtext{To illustrate, we present representative patterns in interaction experience that suggest scaffold updates in the form of rules, tools, and workflows.}

\paragraph{Experience Suggests Adding Rules.}
Agents often overlook domain priors unless explicitly guided.
For example, in DA-Code~\citep{huang2024code}, experience shows that agents often evaluate models using training accuracy without a validation split.
Although the base LLM can conduct proper cross-validation, it will not do so reliably if the scaffold does not emphasize this protocol.
This suggests a simple scaffold update: add a rule that any model evaluation should construct a train/validation split and report performance on the
validation set, rather than on the training data.

\paragraph{Experience Suggests Creating Tools.}
Repetitive or error-prone steps can be replaced by dedicated tools.
For example, experience shows that agents often need to verify whether the solution is non-empty, executable, well-structured, etc.
Creating a tool to run these checks not only simplifies the workflow but also ensures that all these validations are applied.

\paragraph{Experience Suggests Modifying Workflows.}
Failures often stem from the wrong order of actions rather than the lack of capability.
For example, in SWE-bench~\citep{jimenez2023swe}, experience shows that agents often commit changes before submission, leading to an empty patch in evaluation.
A simple update is to add a pre-submission validation step that runs \texttt{git diff --cached} before finish.
Notably, only with this change, an agent can improve the pass-rate by over $2\%$.

\coloredtext{These cases are detailed in Appendix~\ref{Case Studies in Motivation}.}
These examples illustrate that interaction experience reveals agent behavior patterns, which can be leveraged to design better scaffolds.
This motivates an experience-driven scaffold optimizer that inspects execution traces and converts such evidence into targeted scaffold updates.


\section{Method}
In this section, we begin with problem formulation and ReCreate framework, and then detail our agent-as-optimizer design for scaffold optimization.



\subsection{Problem Formulation}
We first formalize the domain agent creation problem.
Given domain $\mathcal{D}$, we denote each task $t_i$ as
\begin{equation}\label{task}
  t_i \triangleq \bigl(x_i, \mathrm{Env}_i\bigr),
\end{equation}
where $x_i$ denotes the problem context and $\mathrm{Env}_i$ denotes an
executable environment of the given domain.
For example, in software engineering, $x$ may contain an issue description and code snippets, while the $\mathrm{Env}$ contains the repository, runtime, and unit tests.

\coloredtext{Given an agent scaffold $\mathcal{A}$}, a base model $\phi$, and task $t_i$, the task agent produces an interaction trajectory
\begin{equation}
  \tau_i \sim P_{\phi}(\cdot \mid \mathcal{A}, t_i),
\end{equation}
where each trajectory $\tau_i$ is a sequence of reasoning steps, tool use, and observations.
After the $\tau_i$ is generated, an agent submission can be $\mathrm{Exec}\bigl(\tau_i, t_i\bigr)$ obtained (e.g., a patch or generated codes).
A task-specific verifier then evaluates this submission and produces a performance metric $r_i$:
\begin{equation}
r_i = \mathrm{Ver}\bigl[\mathrm{Exec}(\tau_i, t_i)\bigr]\in \mathcal{R},
\end{equation}
where $r_i$ could be pass/fail signals from unit tests on software engineering tasks, or scores from evaluation scripts on scientific tasks.

\coloredtext{Given the tuple $(\phi,\mathcal{D},\mathcal{I})$, domain agent creation can then be formulated as the following bi-level optimization problem:
\begin{equation*}
\begin{aligned}
  &\max_{\mathcal{A}}
  \; \mathbb{E}_{t_i \sim \mathcal{D}}
\mathrm{Ver}\bigl[\mathrm{Exec}\bigl(\tau_i^{\ast}(\mathcal{A}, t_i), t_i\bigl)\bigr] \\
  &\text{s.t.}\,
  \tau_i^{\ast}(\mathcal{A}, t_i)
  \in
  \arg\max_{\tau_i \sim P_{\phi}(\cdot \mid \mathcal{A}, t_i)}
  \mathrm{Ver}\bigl(\mathrm{Exec}(\tau_i, t_i)\bigr).
\end{aligned}
\end{equation*}
The inner-level optimization is generating a trajectory $\tau$ to maximize the task performance under current scaffold $\mathcal{A}$.
The outer-level optimization is creating an agent scaffold $\mathcal{A}$ to maximize the expected performance in domain $\mathcal{D}$.}
In practice, the bi-level objective can be approximated via iterative scaffold updates: at iteration $k$, run $\mathcal{A}_k$ on tasks, obtain feedback, and update it to $\mathcal{A}_{k+1}$, starting from $\mathcal{A}_0$ derived from minimal domain information $\mathcal{I}$.

Existing automated agent generation methods can be abstracted as a metric-based update:
\begin{equation*}
\mathcal{A}_{k+1}=\mathrm{Meta}\text{-}\mathrm{Agent}\bigl(\mathcal{A}_k, r\bigr).
\end{equation*}
Here, the entire execution process is compressed into a single metric, which lacks process information.
Instead, we propose to update scaffolds from interaction experience:
\begin{equation}
\begin{aligned}
\mathcal{A}_{k+1}&=\mathrm{ReCreate}\text{-}\mathrm{Agent}\bigl(\mathcal{A}_k, e\bigr), \\
\text{where } \, e &\triangleq \bigl(\tau,\ \mathrm{Exec}, \mathrm{Ver} \bigr).
\end{aligned}
\end{equation}
Here the interaction experience $e$ contains the full trajectory, execution results and evaluation results.
In this way, the outer optimization can use full inner-level information for scaffold updates.

\subsection{The ReCreate Framework}
\label{subsec:recreate}
As illustrated in Figure~\ref{The overview of ReCreate.}, ReCreate formulates the domain agent creation as a bi-level optimization process, which imitates how human experts iteratively refine software systems.
In the inner loop, the agent equipped with scaffold $\mathcal{A}_k$ interacts with the environment to solve tasks.
This loop captures the agent's task-solving process, which contains its task decomposition, chain-of-thought reasoning and tool usage patterns.
In the outer loop, the ReCreate-Agent acts as a scaffold-optimizer.
It inspects the collected experience to attribute why the agent succeeds or fails and generates targeted updates from $\mathcal{A}_k$ to $\mathcal{A}_{k+1}$.
Unlike existing methods that treat agent generation as a black-box optimization problem guided solely by performance metrics, ReCreate reframes it as a white-box debugging process driven by rich interaction experience (i.e., agent trajectories, execution logs, and environmental states).

\begin{figure}[t]
    \centering
    \includegraphics[width=0.95\linewidth]{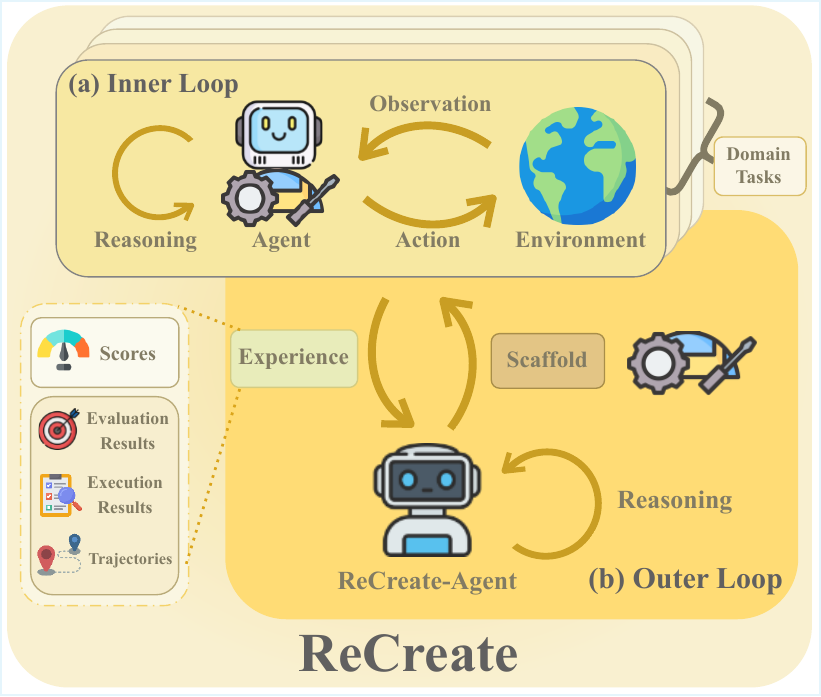}
    \caption{The overview of ReCreate.}
    \label{The overview of ReCreate.}
    \vspace{-1em}
\end{figure}

ReCreate bridges the gap between agent's execution behavior and agent scaffold design, enabling the creation of domain agents from minimal seeds.
Despite its simplicity, the ReCreate framework parallels the workflow of human experts, yet is empowered by superior intelligence.
This embodies a core philosophy: \emph{as models cross the critical threshold of reasoning and creativity, the labor-intensive process of agent creation can finally be automated by the agents themselves.}






\begin{figure*}[htbp]
    \centering
    \includegraphics[width=0.99\linewidth]{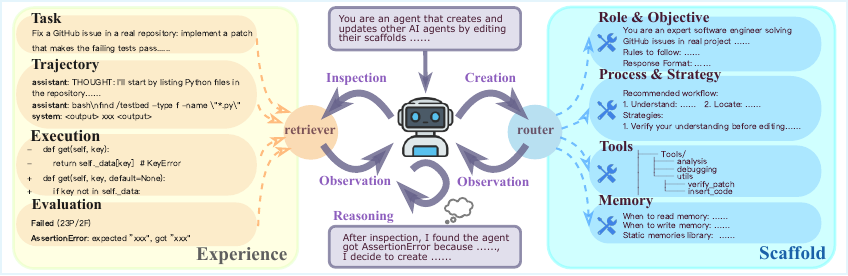}
    \caption{The pipeline of ReCreate. ReCreate-Agent iteratively reasons and acts to locate key evidence on why the agent succeeds or fails and reflect how to improve the agent scaffold.}
    \label{The pipeline of ReCreate.}
    \vspace{-1em}
\end{figure*}

\subsection{The Agent-as-optimizer Design}
\label{The Agent-as-optimizer Design}
While the ReCreate framework leverages interaction experience to improve agent creation, effectively exploiting it is non-trivial for three challenges:
\textbf{(1)}  the full interaction experience is large for LLMs to tackle; \textbf{(2) } attributing agent experience to actionable agent scaffold updates is complex; \textbf{(3)} instance-level fixes often bring overfitting and fail to generalize.
Next, we handle these challenges via three components in the Agent-as-optimizer design.



\paragraph{Experience Storage and Retrieval}
To handle long and noisy experience, we store each task-agent episode as an environment for ReCreate-Agent (call it ReCreate-Environment), which collects the current scaffold, the full interaction trajectories, execution/evaluation results and environment context (e.g., codebase, database and sandbox state).
ReCreate-Environment supports on-demand inspection, allowing the ReCreate-Agent to actively retrieve the relevant piece of experience instead of the full experience.
Typically, the ReCreate-Agent starts from failure or success information and interacts with ReCreate-Environment to progressively narrow down to the key evidence.
To facilitate efficient retrieval, we introduce an \emph{evidence retriever} that indexes critical events (e.g., errors, failing tests, file operations) and links them to their context.
This allows ReCreate-Agent to jump from final evaluation information to the relevant context based on its reasoning capability.

\begin{algorithm}[t]
\caption{ReCreate for Domain Agent Creation}
\label{alg:recreate}
\begin{algorithmic}[1]
\Require LLM $\phi$, dataset $\mathcal{D}$, domain init-info $\mathcal{I}$.
\Ensure final domain scaffold $\mathcal{A}_{\text{final}}$
\State $(\mathcal{D}_{\mathrm{dev}}, \mathcal{D}_{\mathrm{test}}) \gets \textsc{Split}(\mathcal{D})$
\State $\mathcal{A} \gets \textsc{Init}(\mathcal{I})$
\For{$n = 1$ to $N_{\max}$}
  \State $\mathbb{H} \gets \emptyset \,\, $, $\mathcal{B} \gets \textsc{Sample}(\mathcal{D}_{\mathrm{dev}})$
  \For{each task $t \in \mathcal{B}$}
    \State $(\tau, r) \gets \textsc{AgentRun}(\mathcal{A}, \phi, t)$
    \State $(\Delta\mathcal{A}, \kappa) \gets \textsc{Upd}\bigl(\tau, \sigma, \rho, r\bigr)$ \footnotemark 
    \State $\mathbb{H} \gets \mathbb{H} \cup \{(t, \Delta\mathcal{A}, \kappa)\}$
  \EndFor
  \State $\mathcal{A} \gets \textsc{DomUpd}(\mathbb{H}, \mathcal{A}, \mathcal{I})$
\EndFor
\State \Return $\mathcal{A}$ \Comment{report final metrics on $\mathcal{D}_{\mathrm{test}}$}
\end{algorithmic}
\end{algorithm}
\footnotetext{Here, $\sigma$ refers to execution results and $\rho$ refers to evaluation results for short.}

\paragraph{Synergizing Reasoning and Creating}
While the ReCreate-Environment captures comprehensive interaction histories, the raw experiences are often complex to analyze, which brings a gap between experience and agent scaffold creation.
To bridge this gap, the ReCreate-Agent acts as an optimizer for the scaffold in the ReCreate-Agent's environment, as illustrated in Figure~\ref{The pipeline of ReCreate.}.
In the left part, ReCreate-Agent iteratively reasons and inspects the interaction experience to locate key evidence on why the agent succeeds or fails.
The on-demand inspection is enabled by our experience storage and retrieval mechanism described above.
In the right part, ReCreate-Agent iteratively reasons and creates components to improve scaffold.
Specifically, we introduce a \emph{creation router}, including a routing prompt and interfaces for scaffold editing.
The \emph{creation router} guides the ReCreate-Agent to decide \emph{which} scaffold component to edit and \emph{how} to edit it based on the retrieved evidence.
This design ensures that every scaffold update is grounded in specific evidence in the interaction experience, rather than blind trial-and-error.
Based on this pipeline, ReCreate-Agent synergizes Reasoning and Creating for experience-driven agent creation.





\paragraph{Hierarchical Local-to-Domain Updates}
To address the risk of instance-level overfitting, we propose a hierarchical update mechanism, which couples instance-level update $\textsc{Upd}$ with domain-level update $\textsc{DomUpd}$.
At the instance level, the agent analyzes individual interaction experience to generate a candidate update $\Delta\mathcal{A}$ accompanied by its corresponding justification $\kappa$, which are buffered rather than immediately applied.
At the domain level, the ReCreate-Agent synthesizes these instance-level proposals to extract domain patterns.
This hierarchical process filters out task-specific noise, ensuring that only generalized updates are integrated into the final domain agent scaffold.



Algorithm~\ref{alg:recreate} summarizes the complete workflow of ReCreate.
First, the domain dataset $\mathcal{D}$ is split into development set $\mathcal{D}_{\mathrm{dev}}$ and test set $\mathcal{D}_{\mathrm{test}}$, where $\mathcal{D}_{\mathrm{dev}}$ is used for agent scaffold creation and $\mathcal{D}_{\mathrm{test}}$ is used for agent scaffold evaluation.
The agent scaffold $\mathcal{A}$ is initialized through minimal initial information $\mathcal{I}$, including environment interfaces and necessary procedures.
For each task in the sampled batch $\mathcal{B}$, ReCreate-Agent derives a local update proposal $\Delta\mathcal{A}$ from interaction experience and buffers it into $\mathbb{H}$ (Lines 5--8).
These buffered local update proposals are aggregated to global update by ReCreate Agent (Line 10).
Finally, the created agent scaffold $\mathcal{A}$ is evaluated on $\mathcal{D}_{\mathrm{test}}$.

\subsection{Comparing with Existing Methods}
\paragraph{Comparison to Existing Self-Evolve Methods.}
Recent self-evolving methods~\citep{xia2025live, yang2023large, zhao2024expel} also leverage experience to refine pre‑existing agents. ReCreate differs from self-evolving methods in three aspects:
(1) \textbf{Scope}: Instead of refining pre-defined scaffolds, ReCreate builds agents from scratch, broadening applicability to scenarios without mature agents. (2) \textbf{Objective}: Unlike these methods prioritizing instance-level success, ReCreate aims for domain-level generalization with hierarchical updates. (3) \textbf{Strategy}: Rather than relying on high-level outcomes, ReCreate conducts  \emph{fine-grained inspection} of execution traces, extracting concrete meaningful evidence for optimization. 
Empirically, ReCreate even initialized with a minimal scaffold outperforms these methods with fully-developed scaffolds (\cf Section 5).
Beyond this, we provide a detailed discussion in Appendix~\ref{sec:discussion} about the design of our Agent-as-optimizer and how ReCreate differs from existing methods.


\begin{table*}[t]
\centering
\setlength{\tabcolsep}{5pt}
\renewcommand{\arraystretch}{1.1}
{\fontsize{9}{11}\selectfont
\begin{tabular}{ll|ccc|ccc|cc|c}
\toprule
\multirow{2}{*}{\textbf{Domain}} & \multirow{2}{*}{\textbf{Dataset}} &
\multicolumn{3}{c|}{\textbf{Human-designed}} &
\multicolumn{3}{c|}{\textbf{Self-Evolve}} &
\multicolumn{2}{c|}{\textbf{Agent Generation}} &
\multicolumn{1}{c}{\cellcolor{red!12}{\textbf{Ours}}} \\
& &
\textbf{CoT} & \textbf{SBA} & \textbf{Refine} &
\textbf{LIVE} & \textbf{OPRO} & \textbf{ExpeL} &
\textbf{ADAS} & \textbf{Square} &
\cellcolor{red!12}{\textbf{ReCreate}} \\
\toprule
\multirow{2}{*}{SWE}
& \emph{Django}      & $58.29$ & $58.77$ & $56.87$ & $58.77$ & $52.13$ & $59.24$ & $55.92$ & $57.82$ & \cellcolor{red!12}{$\mathbf{60.19}$} \\
& \emph{Sympy}       & $61.82$ & $61.82$ & $58.18$ & $58.18$ & $54.55$ & $56.36$ & $58.18$ & $54.55$ & \cellcolor{red!12}{$\mathbf{63.64}$} \\
\hline
\multirow{6}{*}{DS}
& \emph{DW} & $42.81$ & $41.66$ & $42.43$ & $38.55$ & $18.55$ & $47.50$ & $37.98$ & $45.98$ & \cellcolor{red!12}{$\mathbf{51.94}$} \\
& \emph{ML} & $34.32$ & $36.25$ & $35.85$ & $41.68$ & $39.08$ & $40.27$ & $21.53$ & $22.90$ & \cellcolor{red!12}{$\mathbf{42.88}$} \\
& \emph{SA} & $19.33$ & $20.15$ & $20.39$ & $21.83$ & $17.16$ & $22.44$ & $15.66$ & $11.99$ & \cellcolor{red!12}{$\mathbf{24.50}$} \\
& \emph{Numpy}       & $62.00$ & $64.50$ & $64.00$ & $68.00$ & $71.00$ & $68.50$ & $61.00$ & $67.00$ & \cellcolor{red!12}{$\mathbf{77.00}$} \\
& \emph{Pandas}      & $62.73$ & $61.25$ & $63.10$ & $64.21$ & $63.47$ & $64.94$ & $60.52$ & $63.10$ & \cellcolor{red!12}{$\mathbf{68.63}$} \\
& \emph{Matplotlib}  & $78.52$ & $80.74$ & $81.48$ & $82.96$ & $82.96$ & $78.52$ & $76.30$ & $82.96$ & \cellcolor{red!12}{$\mathbf{85.19}$} \\
\hline
\multirow{3}{*}{Math}
& \emph{Algebra}     & $81.45$ & $85.48$ & $84.68$ & $85.48$ & $87.10$ & $83.87$ & $80.65$ & $83.87$ & \cellcolor{red!12}{$\mathbf{92.74}$} \\
& \emph{NT}         & $91.94$ & $90.32$ & $90.32$ & $93.55$ & $\mathbf{100.00}$ & $90.32$ & $83.87$ & $93.55$ & \cellcolor{red!12}{$\mathbf{100.00}$} \\
& \emph{C\&P}         & $94.74$ & $94.74$ & $94.74$ & $92.11$ & $94.74$ & $92.11$ & $84.21$ & $94.74$ & \cellcolor{red!12}{$\mathbf{100.00}$} \\
\hline
\multirow{2}{*}{Digital}
& \emph{Normal}      & $48.81$ & $48.81$ & $47.02$ & $47.62$ & $51.79$ & $49.40$ & $50.00$ & $48.81$ & \cellcolor{red!12}{{$\mathbf{52.98}$}} \\
& \emph{Challenge}   & $34.05$ & $36.21$ & $35.01$ & $34.53$ & $34.29$ & $34.53$ & $36.93$ & $34.77$ & \cellcolor{red!12}{$\mathbf{40.29}$} \\
\hline
\multicolumn{2}{c|}{\textbf{Average}} & $59.29$ & $60.05$ & $59.54$ & $60.57$ & $58.99$ & $60.62$ & $55.60$ & $58.62$ & \cellcolor{red!12}{$\mathbf{66.15}$} \\
\bottomrule
\end{tabular}
}
\vspace{-0.1cm}
\caption{Pass rate or testing score on various real-world agent benchmarks.}
\vspace{-0.2em}
\label{main_results}
\end{table*}

\section{Experiments and Results}
In this section, we evaluate ReCreate from the following perspectives: (1) comparison against baselines on thirteen benchmarks across four domains; (2) behavioral analysis of the ReCreate-Agent; (3) multi-level ablation studies; (4) analysis of the update dynamics and cost-effectiveness.

\subsection{Experimental Setup}
\paragraph{Datasets}
To validate the effectiveness of ReCreate across diverse real-world scenarios, we conduct experiments on four representative domains widely used for agent evaluation, including Software Engineering (SWE), Data Science (DS), Mathematics (Math), and Digital Assistance (Digital).
Specifically, we instantiate these domains by using their most representative subsets: 
for SWE, we select the two largest repositories, \emph{Django} and \emph{SymPy}, from the SWE-bench-Verified~\citep{jimenez2023swe};
for DS, we select the three largest subsets from DA-Code~\citep{huang2024code} (\emph{Data Wrangling}, \emph{Machine Learning}, \emph{Statistical Analysis}) and DS-1000~\citep{lai2023ds} (\emph{NumPy}, \emph{Pandas}, \emph{Matplotlib});
for Math, we select the three sub-domains in MATH~\citep{hendrycks2021measuring} (\emph{Algebra}, \emph{Number Theory}, \emph{Counting\&Probability});
for Digital, we select both the \emph{Normal} and \emph{Challenge} subsets of AppWorld~\citep{trivedi2024appworld}.
It is worth noting that DA-Code uses a continuous evaluation signal, producing a score in $[0,1]$ (converted to a percentage in our reporting). In contrast, all other benchmarks considered here use deterministic binary outcomes ($0$ for failure and $1$ for success).
Detailed information for datasets is shown in Appendix~\ref{Detailed Information for Datasets}.

\begin{figure*}[t]
    \centering
    \begin{subfigure}[t]{0.3\textwidth}
        \centering
        \includegraphics[width=\linewidth]{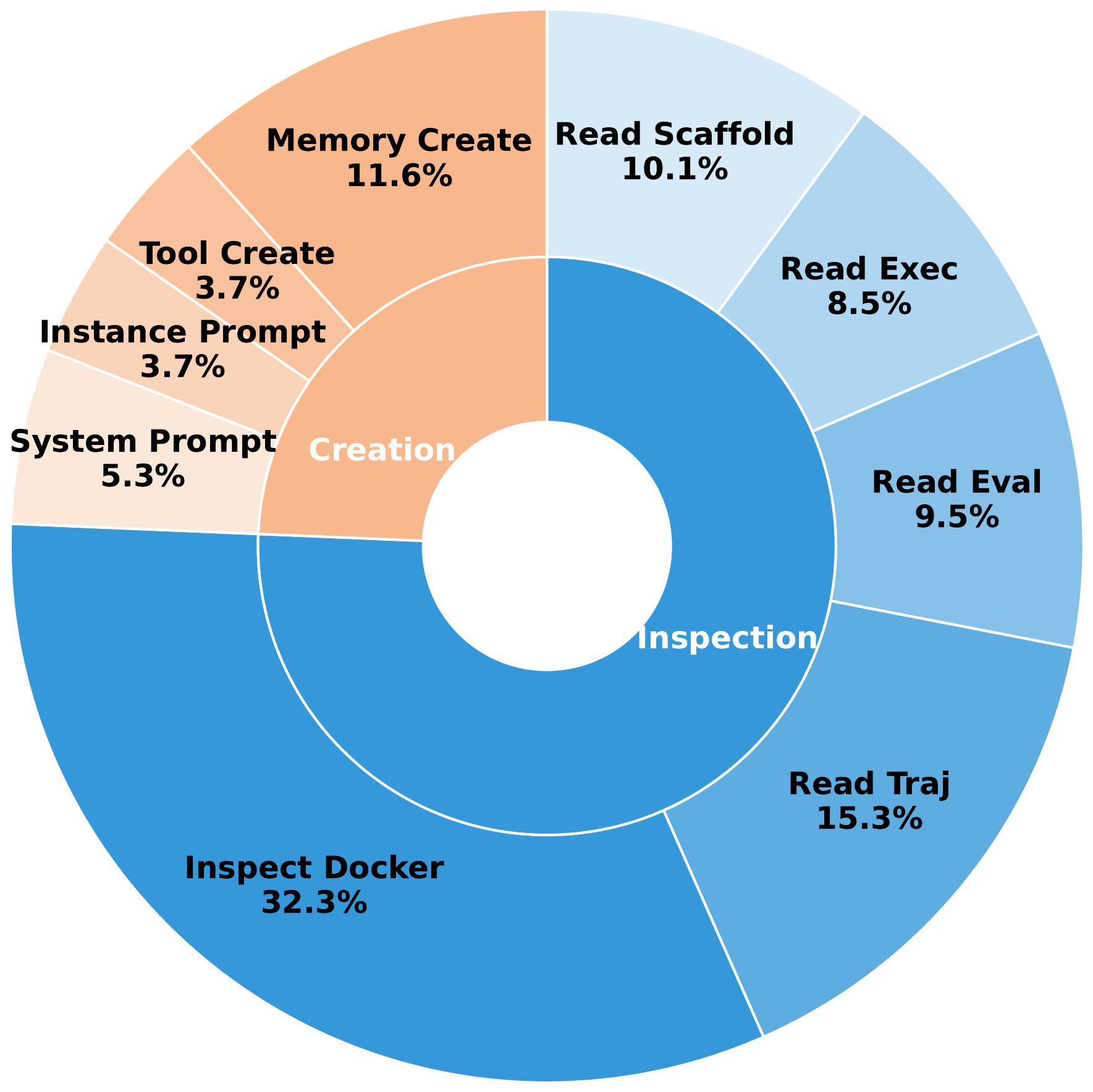}
        \caption{Django}
        \label{fig:sunburst-django}
    \end{subfigure}\hfill
    \begin{subfigure}[t]{0.3\textwidth}
        \centering
        \includegraphics[width=\linewidth]{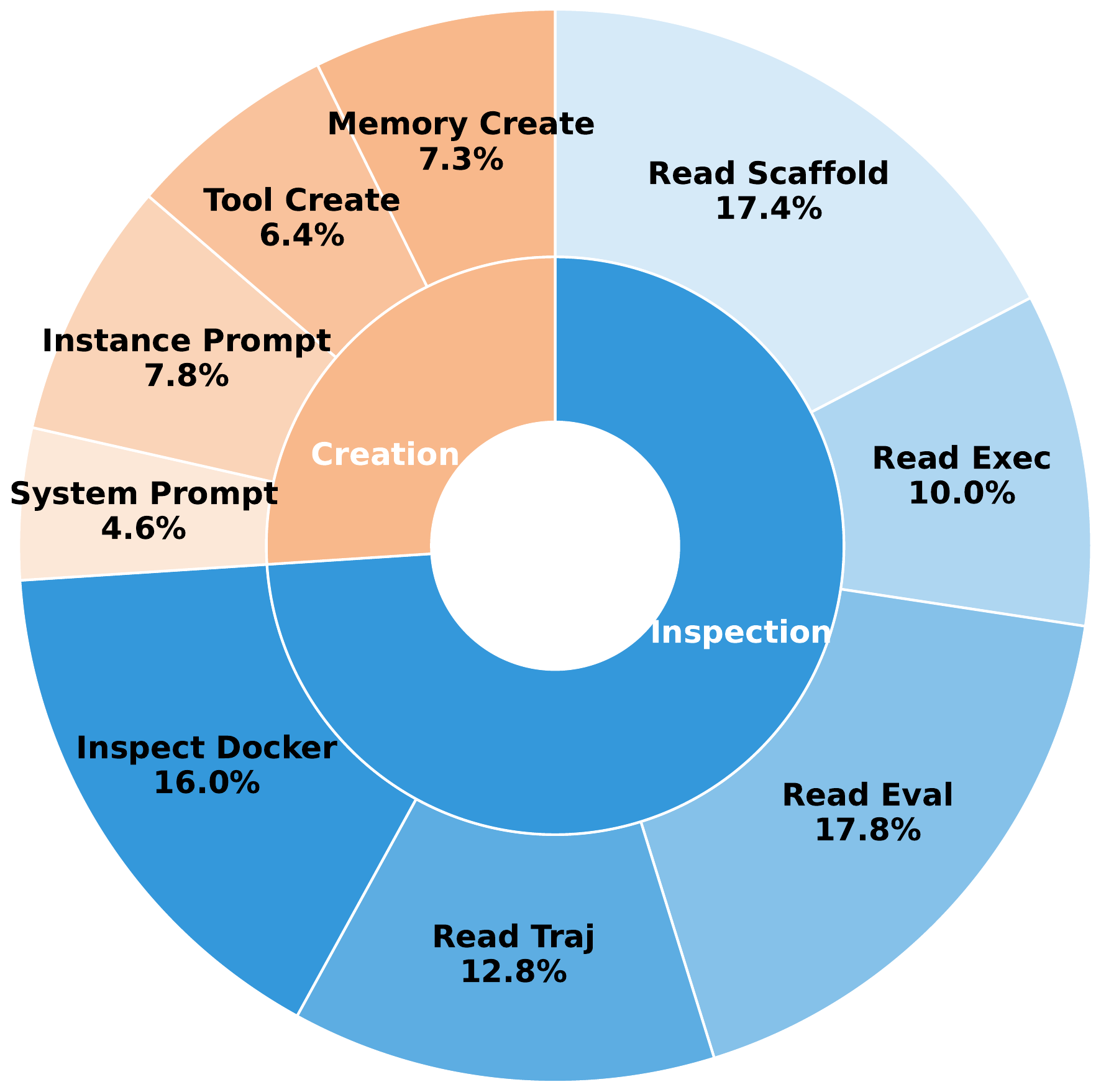}
        \caption{Machine Learning}
        \label{fig:sunburst-2}
    \end{subfigure}\hfill
    \begin{subfigure}[t]{0.3\textwidth}
        \centering
        \includegraphics[width=\linewidth]{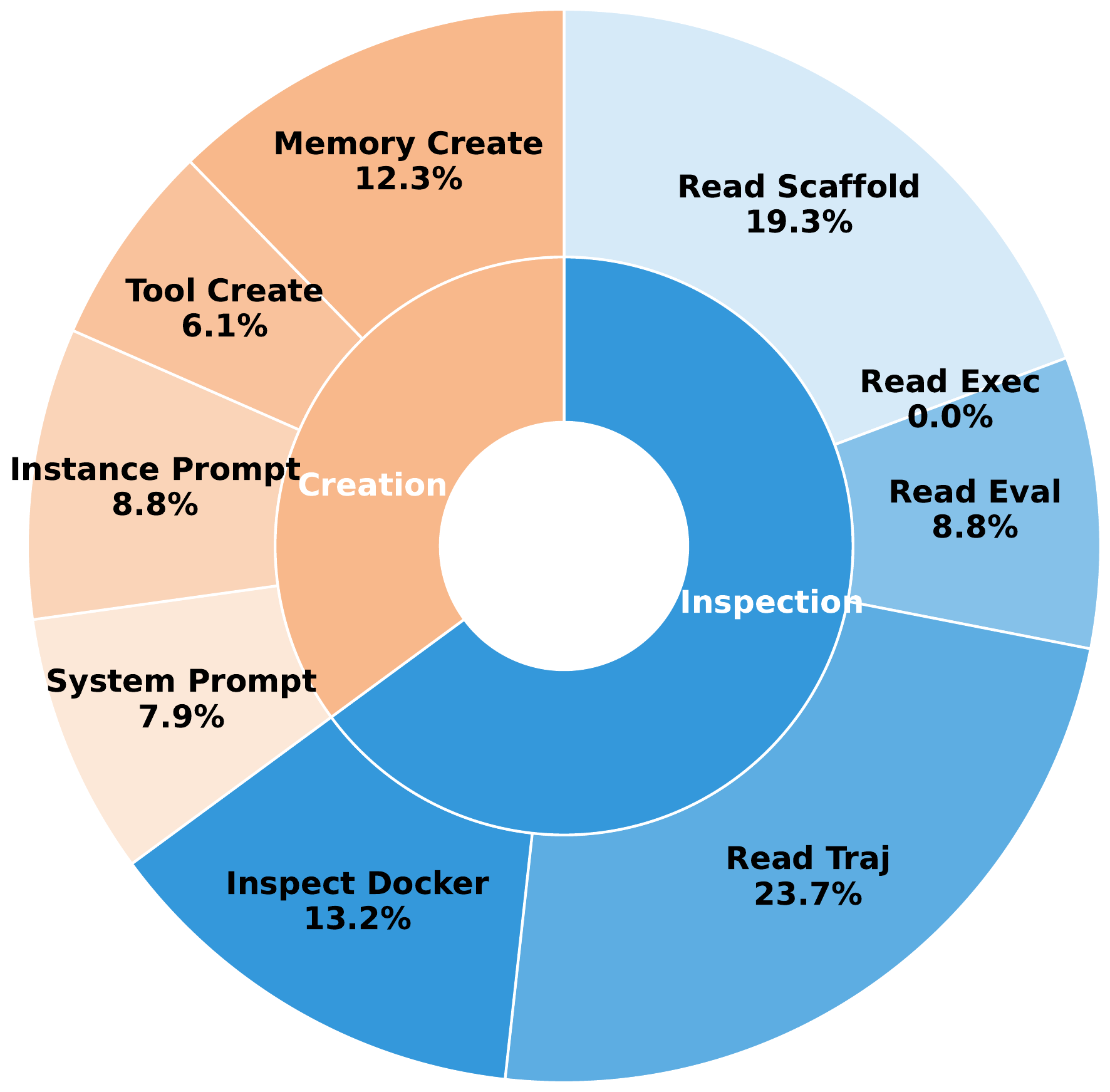}
        \caption{AppWorld}
        \label{fig:sunburst-3}
    \end{subfigure}
    \vspace{-0.2em}
    \caption{Action distributions of the ReCreate-Agent in various domains.}
    \label{Action distributions of the ReCreate-Agent.}
    \vspace{-1em}
\end{figure*}

\paragraph{Implementations}
We employ \texttt{gpt-5-mini} as the backbone for the task agent to ensure inference efficiency and employ \texttt{claude-4.5-opus} as the ReCreate-Agent to guarantee high-quality reasoning and scaffold updates.
We set the temperature to $0$ for the \texttt{claude-4.5-opus} and $1.0$ for the \texttt{gpt-5-mini} \coloredtext{(fixed at 1.0 by the API)}.
Following Algorithm~\ref{alg:recreate}, we configure the loop with a maximum iteration $N_{\max}=2$ and a batch size of $4$.
For data partitioning, we randomly sample a small set of approximately $20$ instances as the development set $\mathcal{D}_{\mathrm{dev}}$ for each domain, reserving all remaining data for testing \coloredtext{(ranging from $38$ to $417$)}.
All tasks are executed within Docker sandboxes.
Detailed statistics of data splits and prompts for the ReCreate-Agent are provided in Appendix~\ref{Detailed Information for Datasets}.

\paragraph{Baselines}
We compare ReCreate against three related categories of methods to ensure a comprehensive evaluation.
The first category is human-designed scaffolds with test-time scaling, including CoT~\citep{wei2022chain}, Step-Back Abstraction (short as SBA)~\citep{zheng2023take}, and Self-Refine (short as Refine)~\citep{madaan2023self}.
The second category is Self-Evolving methods, where agents autonomously refine themselves for solving tasks.
We select representative methods for different evolution targets: LIVE~\citep{xia2025live} for tool evolution, OPRO~\citep{yang2023large} for prompt optimization, and ExpeL~\citep{zhao2024expel} for experience accumulation.
The third category is Automated Agent Generation, including ADAS~\citep{hu2024automated}, AgentSquare~\citep{shang2024agentsquare} (short as Square).


\subsection{Main Results}
Table~\ref{main_results} reports the main results across four domains.
First, ReCreate consistently exceeds both human-designed scaffolds and self-evolving baselines across all domains.
On average across all benchmarks, ReCreate improves the overall score by more than $5\%$ over the strongest competing method, especially with clear performance gains on DS, Math, and Digital tasks.
These notable results are  because these baselines rely on human prior knowledge encoded in hand-crafted scaffolds, which can be difficult to acquire and may not generalize well when domain knowledge is scarce.
Second, ReCreate delivers substantial gains over Agent Generation methods, improving the overall average by more than $7\%$.
\coloredtext{This highlights the effectiveness of leveraging interaction experience, rather than relying solely on a scalar score, for domain agent creation.
Besides, Agent Generation methods typically search for or compose agents from a pre-built component pool, while ReCreate updates the scaffold directly from execution experience, without requiring any predefined modules.}


\subsection{Statistical Study}
To look into the creation process, we count the action distribution of the ReCreate-Agent in three sub-domains, shown in
Figure~\ref{Action distributions of the ReCreate-Agent.}.
Across all cases, inspection dominates creation (roughly 65\%--76\% vs.\ 24\%--35\%), indicating that ReCreate-Agent typically \emph{locates and verifies evidence} more than proposing scaffold edits.
This aligns with our agent-as-optimizer design, where scaffold updates are grounded in execution experience rather than blind trial-and-error.

The dominant evidence sources and update targets vary by domain.
On Django, the agent heavily inspects the Docker sandbox for code base, while creation mainly manifests as memory construction to consolidate debugging findings into reusable rules.
On Machine Learning, inspection frequently focuses on scaffolds and evaluation artifacts, and creation is more balanced across prompt adjustments and tool/memory creations.
On AppWorld, trajectory inspection is prominent and creation becomes notably more frequent, with more prompt and memory updates.
In short, the agent's behavior is highly context-dependent, allocating its reasoning and creation efforts where they yield the highest value for the specific task.

\begin{figure}[t]
\vspace{-0.5em}
    \centering
    \includegraphics[width=0.98\linewidth]{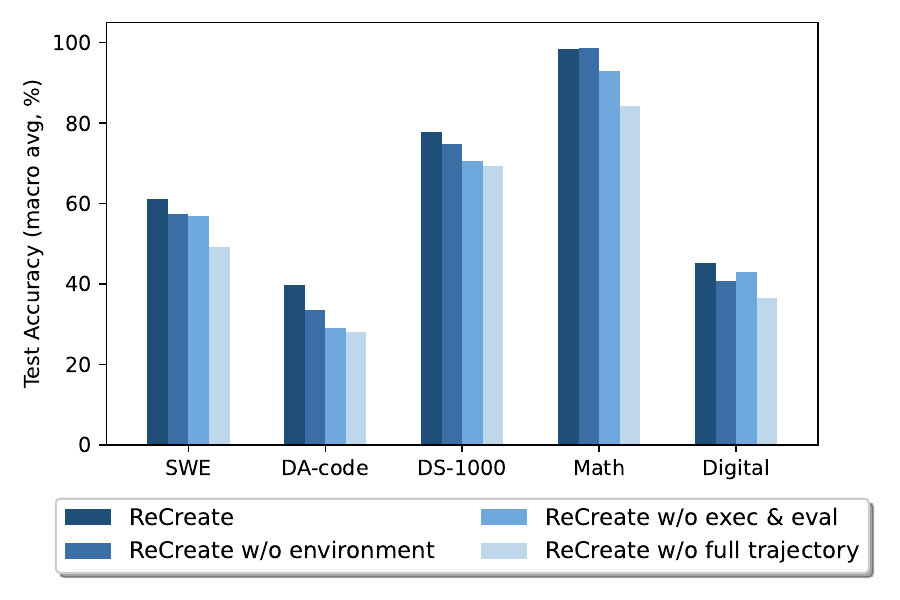}
    \vspace{-0.5em}
    \caption{Ablations on experience components.}
    \label{Ablations on experience components.}
    \vspace{-0.5em}
\end{figure}

\subsection{Ablation Study}

\paragraph{Observation-level Ablation}
Figure~\ref{Ablations on experience components.} reports ablations over the experience components in ReCreate.
Across all five domains, the full ReCreate model has the best performance. Removing any component degrades its performance, which confirms the importance of experience components.
Among the variants, removing the full trajectory causes the largest and most consistent performance drop, highlighting that step-by-step traces provide crucial context for diagnosing failures and guiding creation.
Removing execution \& evaluation feedback also leads to the performance drop, suggesting that outcome signals (e.g., generated files, test results, verifier feedback) are necessary to anchor updates.
Removing the environment yields a smaller but consistent decline, indicating the value of runnable execution for faithful inspection and debugging.
These results underscore the complementary roles of trajectory, environment and exec/eval feedback for domain agent creation.

\begin{table}[t]
\centering
\setlength{\tabcolsep}{3pt}
\renewcommand{\arraystretch}{1}
\begin{tabular}{lccc}
\toprule
\textbf{Method} & \textbf{SWE} & \textbf{DA-Code} & \textbf{DS}  \\
\midrule
\rowcolor{red!12} ReCreate & 60.19 & 39.77 & 77.74  \\
\hspace{0.4em} w/o creation router & 58.00 & 37.13 & 75.39  \\
\hspace{0.4em} w/o $\textsc{DomUpd}$ & 57.09 & 37.83 & 75.96 \\
\bottomrule
\end{tabular}
\vspace{-0.5em}
\caption{Action-level ablation.}
\label{action-ablation}
\vspace{-0.2em}
\end{table}

\paragraph{Action-level Ablation}
Table~\ref{action-ablation} ablates two action-level components in ReCreate.
Removing either creation router or domain update $\textsc{DomUpd}$ consistently hurts the performance across SWE, DA-Code, and DS-1000 (DS for short).
Without \emph{creation router}, the ReCreate-Agent tends to focus on instance prompt; without $\textsc{DomUpd}$, the updates are biased toward instance details.
Therefore, the two components are complementary: creation router improves execution reliability of ReCreate-Agent, while $\textsc{DomUpd}$ improves cross-task generalization.

\begin{figure}[t]
\vspace{-0.2em}
    \centering
    \includegraphics[width=0.85\linewidth]{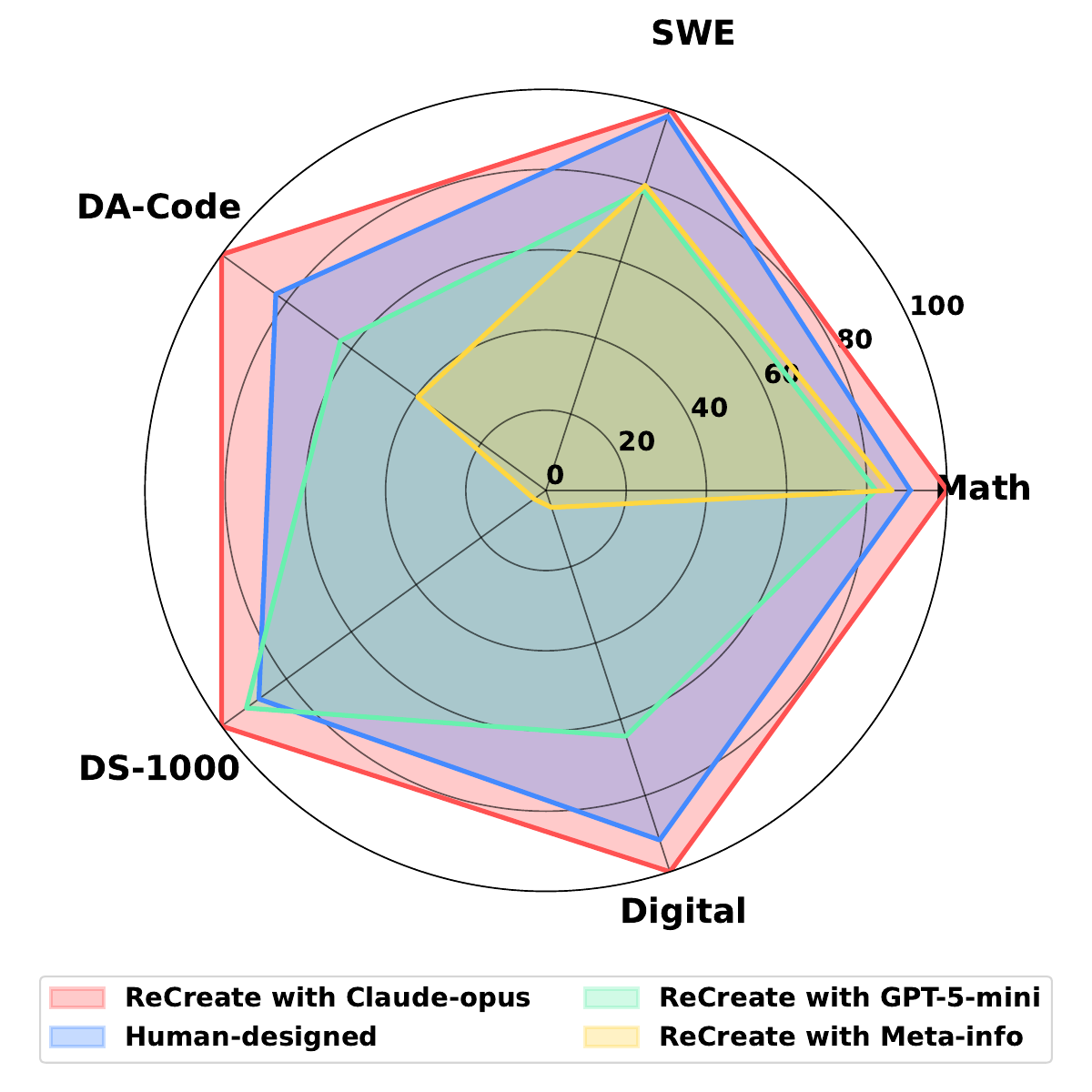}
    \vspace{-0.5em}
    \caption{Reasoning-level ablation.}
    \label{Reasoning-level Ablation.}
    \vspace{-1.2em}
\end{figure}

\paragraph{Reasoning-level Ablation}
The ReCreate framework requires strong reasoning capability to achieve reliable agent creation.
Figure~\ref{Reasoning-level Ablation.} compares ReCreate-Agent with different reasoning capacities and the task agent is fixed as \texttt{gpt-5-mini}.
The radar values are normalized by ReCreate with \texttt{Claude-opus}, indicating each setting's relative performance ratio.
We draw two conclusions.
First, ReCreate with only initial domain information yields very poor performance in most domains (except Math).
This indicates that initial domain  information alone is far from sufficient and that effective scaffolds require richer domain knowledge from interaction experience or experts.
Second, ReCreate with \texttt{Claude-opus} consistently surpasses \texttt{Human-designed} scaffolds, whereas ReCreate with \texttt{gpt-5-mini} fails to outperform them in most domains.
This gap indicates that the stronger reasoning capability substantially improves the ReCreate-Agent's ability to interpret interaction experience and translate it into actionable scaffold updates.
Furthermore, it suggests that frontier LLMs are approaching the point of matching or even replacing expert-designed scaffolds in practice.

\subsection{Analysis of the Update Procedure}
\begin{figure}[t]
    \centering
    \includegraphics[width=0.98\linewidth]{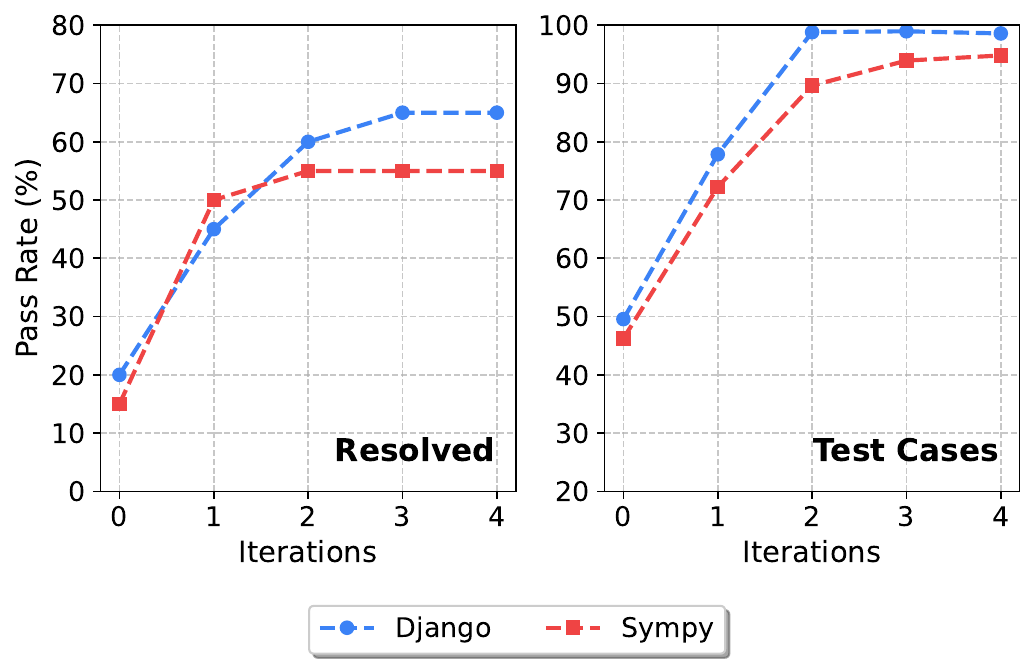}
    \vspace{-0.5em}
    \caption{Post-update Gain Verification on SWE tasks.}
    \label{Post-update Gain Verification on SWE tasks.}
    \vspace{-1.2em}
\end{figure}

In this section, we dive into the update process in ReCreate and study how it affects agent performance on dev-set $\mathcal{D}_{\mathrm{dev}}$ and test set $\mathcal{D}_{\mathrm{test}}$.
We consider a single-case setting where the development set contains only one instance. 
Starting from the initial scaffold, we iteratively apply ReCreate updates and re-evaluate the agent on the same case to quantify the post-update gain in pass rate. 
Figure~\ref{Post-update Gain Verification on SWE tasks.} plots the average pass rate curve for $20$ single-case development sets (i.e., $20$ distinct cases) in the SWE domain under this setting, with a maximum of $N_{\max}=4$ update iterations.
The left panel shows the average task-level resolved rate, while the right panel reports the fine-grained test-case pass rate.
Under this setting, ReCreate steadily improves both the task’s resolved rate and its test-case pass rate by iteratively updating the agent scaffold.
These results suggest that ReCreate can fix a subset of tasks that initially failed through its experience-driven updates on agent scaffold edits.

\subsection{Case Study}
We present task-solving cases to intuitively showcase how ReCreate evolves the agent scaffold; details are provided in Appendix~\ref{Case Study}.

\subsection{Cost}
\begin{figure}
    \centering
    \includegraphics[width=0.98\linewidth]{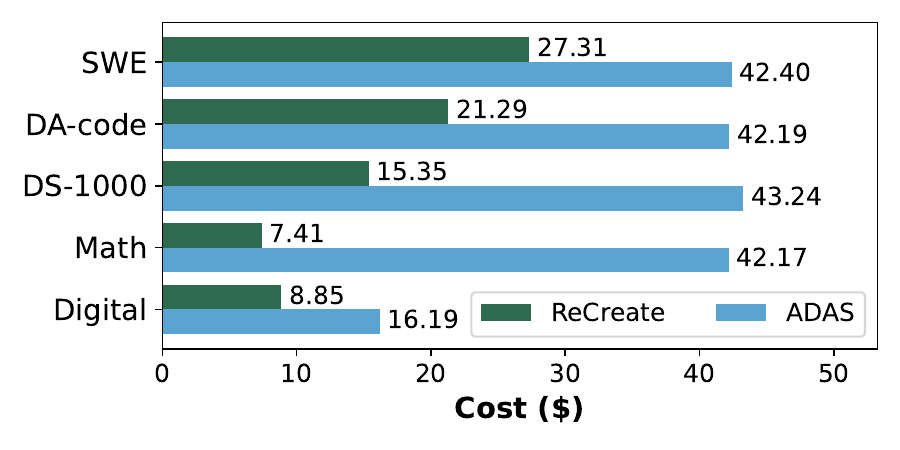}
    \vspace{-1em}
    \caption{Cost Comparison.}
    \label{Cost Comparison.}
    \vspace{-1em}
\end{figure}

Beyond performance, we also assessed the cost-effectiveness of ReCreate compared to automated agent generation methods.
Figure~\ref{Cost Comparison.} compares the average cost (counted by USD) of scaffold optimization across domains between ReCreate and ADAS.
ReCreate is more efficient than ADAS, reducing the cost by roughly $36\%$ to $82\%$.
Even though ReCreate employs a strong ReCreate-Agent (e.g., \texttt{claude-4.5-opus}) for scaffold updates, it converges with a small development set and fewer iterations thanks to the rich signals from interaction experience.
In contrast, ADAS has to repeatedly evaluate each candidate, leading to higher cost.

\section{Conclusion}
We introduced {ReCreate}, an experience-driven framework for domain agent creation that optimizes agent scaffolds by learning from interaction experience rather than relying solely on performance metrics. Concretely, {ReCreate} adopts an agent-as-optimizer design with three components, enabling scaffold updates grounded in concrete evidence while improving task generalization.
Empirically, {ReCreate} yields consistent performance gains over baselines across diverse domains even when starting from minimal seed scaffolds.  


\section{Limitations}

The limitations of this work are twofold.
First, ReCreate focuses on optimizing agent scaffolds at the textual and code levels, such as prompts, reasoning strategies, and tool implementations. It does not extend to infrastructure adaptations, such as harness and environments, as these require heavy engineering distinct from the generalizable logic of agent creation.
Second, ReCreate does not update base model parameters. Combining the discovered scaffold patterns with model fine-tuning is a promising but computationally expensive direction for future work.




\section{Acknowledgements}
This work is supported by the Zhejiang Province “JianBingLingYan+X” Research and Development Plan (2025C02020).
We also sincerely thank Yi Liu and the CodeBuddy Team in Tencent CodeBuddy (https://www.codebuddy.ai/) for their valuable assistance throughout this work.







\bibliography{custom}

@misc{gpt52,
  author       = {{OpenAI}},
  title        = {Introducing GPT-5.2},
  year         = {2025},
  howpublished = {\url{https://openai.com/index/introducing-gpt-5-2}}
}

@misc{Raising,
  author       = {{Anthropic}},
  title        = {Raising the bar on SWE-bench Verified with Claude 3.5 Sonnet},
  year         = {2025},
  howpublished = {\url{https://www.anthropic.com/engineering/swe-bench-sonnet}}
}

@misc{gemini3,
  author       = {{Google}},
  title        = {A new era of intelligence with Gemini 3},
  year         = {2025},
  howpublished = {\url{https://blog.google/products/gemini/gemini-3/}}
}

@misc{skills,
  author       = {{Han Lee}},
  title        = {Claude Agent Skills: A First Principles Deep Dive},
  year         = {2025},
  howpublished = {\url{https://leehanchung.github.io/blogs/2025/10/26/claude-skills-deep-dive/}}
}

@misc{anthropic2025claude45,
  author       = {{Anthropic}},
  title        = {Introducing Claude Opus 4.5},
  year         = {2025},
  howpublished = {\url{https://www.anthropic.com/news/claude-opus-4-5}}
}

@article{liu2025deepseek,
  title={DeepSeek-V3. 2: Pushing the Frontier of Open Large Language Models},
  author={Liu, Aixin and Mei, Aoxue and Lin, Bangcai and Xue, Bing and Wang, Bingxuan and Xu, Bingzheng and Wu, Bochao and Zhang, Bowei and Lin, Chaofan and Dong, Chen and others},
  journal={arXiv preprint arXiv:2512.02556},
  year={2025}
}

@article{luo2025large,
  title={Large language model agent: A survey on methodology, applications and challenges},
  author={Luo, Junyu and Zhang, Weizhi and Yuan, Ye and Zhao, Yusheng and Yang, Junwei and Gu, Yiyang and Wu, Bohan and Chen, Binqi and Qiao, Ziyue and Long, Qingqing and others},
  journal={arXiv preprint arXiv:2503.21460},
  year={2025}
}

@article{xi2025rise,
  title={The rise and potential of large language model based agents: A survey},
  author={Xi, Zhiheng and Chen, Wenxiang and Guo, Xin and He, Wei and Ding, Yiwen and Hong, Boyang and Zhang, Ming and Wang, Junzhe and Jin, Senjie and Zhou, Enyu and others},
  journal={Science China Information Sciences},
  volume={68},
  number={2},
  pages={121101},
  year={2025},
  publisher={Springer}
}

@inproceedings{wang2024empowering,
  title={Empowering large language models: Tool learning for real-world interaction},
  author={Wang, Hongru and Qin, Yujia and Lin, Yankai and Pan, Jeff Z and Wong, Kam-Fai},
  booktitle={Proceedings of the 47th International ACM SIGIR Conference on Research and Development in Information Retrieval},
  pages={2983--2986},
  year={2024}
}

@article{suzgun2024meta,
  title={Meta-prompting: Enhancing language models with task-agnostic scaffolding},
  author={Suzgun, Mirac and Kalai, Adam Tauman},
  journal={arXiv preprint arXiv:2401.12954},
  year={2024}
}

@article{li2025deepagent,
  title={DeepAgent: A General Reasoning Agent with Scalable Toolsets},
  author={Li, Xiaoxi and Jiao, Wenxiang and Jin, Jiarui and Dong, Guanting and Jin, Jiajie and Wang, Yinuo and Wang, Hao and Zhu, Yutao and Wen, Ji-Rong and Lu, Yuan and others},
  journal={arXiv preprint arXiv:2510.21618},
  year={2025}
}

@article{jimenez2023swe,
  title={Swe-bench: Can language models resolve real-world github issues?},
  author={Jimenez, Carlos E and Yang, John and Wettig, Alexander and Yao, Shunyu and Pei, Kexin and Press, Ofir and Narasimhan, Karthik},
  journal={arXiv preprint arXiv:2310.06770},
  year={2023}
}

@article{yang2025code,
  title={From Code Foundation Models to Agents and Applications: A Practical Guide to Code Intelligence},
  author={Yang, Jian and Zhang, Wei and Liu, Shark and Wu, Jiajun and Guo, Shawn and Li, Yizhi},
  journal={arXiv preprint arXiv:2511.18538},
  year={2025}
}

@article{yang2024swe,
  title={Swe-agent: Agent-computer interfaces enable automated software engineering},
  author={Yang, John and Jimenez, Carlos E and Wettig, Alexander and Lieret, Kilian and Yao, Shunyu and Narasimhan, Karthik and Press, Ofir},
  journal={Advances in Neural Information Processing Systems},
  volume={37},
  pages={50528--50652},
  year={2024}
}

@article{tang2025ai,
  title={AI-Researcher: Autonomous Scientific Innovation},
  author={Tang, Jiabin and Xia, Lianghao and Li, Zhonghang and Huang, Chao},
  journal={arXiv preprint arXiv:2505.18705},
  year={2025}
}

@article{weng2024cycleresearcher,
  title={Cycleresearcher: Improving automated research via automated review},
  author={Weng, Yixuan and Zhu, Minjun and Bao, Guangsheng and Zhang, Hongbo and Wang, Jindong and Zhang, Yue and Yang, Linyi},
  journal={arXiv preprint arXiv:2411.00816},
  year={2024}
}

@article{team2025tongyi,
  title={Tongyi deepresearch technical report},
  author={Team, Tongyi DeepResearch and Li, Baixuan and Zhang, Bo and Zhang, Dingchu and Huang, Fei and Li, Guangyu and Chen, Guoxin and Yin, Huifeng and Wu, Jialong and Zhou, Jingren and others},
  journal={arXiv preprint arXiv:2510.24701},
  year={2025}
}

@article{he2024webvoyager,
  title={Webvoyager: Building an end-to-end web agent with large multimodal models},
  author={He, Hongliang and Yao, Wenlin and Ma, Kaixin and Yu, Wenhao and Dai, Yong and Zhang, Hongming and Lan, Zhenzhong and Yu, Dong},
  journal={arXiv preprint arXiv:2401.13919},
  year={2024}
}

@article{wang2024openhands,
  title={Openhands: An open platform for ai software developers as generalist agents},
  author={Wang, Xingyao and Li, Boxuan and Song, Yufan and Xu, Frank F and Tang, Xiangru and Zhuge, Mingchen and Pan, Jiayi and Song, Yueqi and Li, Bowen and Singh, Jaskirat and others},
  journal={arXiv preprint arXiv:2407.16741},
  year={2024}
}

@misc{openmanus2025,
  author = {Xinbin Liang and Jinyu Xiang and Zhaoyang Yu and Jiayi Zhang and Sirui Hong and Sheng Fan and Xiao Tang},
  title = {OpenManus: An open-source framework for building general AI agents},
  year = {2025},
  publisher = {Zenodo},
  doi = {10.5281/zenodo.15186407},
  url = {https://doi.org/10.5281/zenodo.15186407},
}

@article{xia2024agentless,
  title={Agentless: Demystifying llm-based software engineering agents},
  author={Xia, Chunqiu Steven and Deng, Yinlin and Dunn, Soren and Zhang, Lingming},
  journal={arXiv preprint arXiv:2407.01489},
  year={2024}
}

@article{li2025search,
  title={Search-o1: Agentic search-enhanced large reasoning models},
  author={Li, Xiaoxi and Dong, Guanting and Jin, Jiajie and Zhang, Yuyao and Zhou, Yujia and Zhu, Yutao and Zhang, Peitian and Dou, Zhicheng},
  journal={arXiv preprint arXiv:2501.05366},
  year={2025}
}

@article{ma2024sciagent,
  title={Sciagent: Tool-augmented language models for scientific reasoning},
  author={Ma, Yubo and Gou, Zhibin and Hao, Junheng and Xu, Ruochen and Wang, Shuohang and Pan, Liangming and Yang, Yujiu and Cao, Yixin and Sun, Aixin and Awadalla, Hany and others},
  journal={arXiv preprint arXiv:2402.11451},
  year={2024}
}

@article{li2025codepde,
  title={CodePDE: An Inference Framework for LLM-driven PDE Solver Generation},
  author={Li, Shanda and Marwah, Tanya and Shen, Junhong and Sun, Weiwei and Risteski, Andrej and Yang, Yiming and Talwalkar, Ameet},
  journal={arXiv preprint arXiv:2505.08783},
  year={2025}
}

@article{hu2024automated,
  title={Automated design of agentic systems},
  author={Hu, Shengran and Lu, Cong and Clune, Jeff},
  journal={arXiv preprint arXiv:2408.08435},
  year={2024}
}

@article{shang2024agentsquare,
  title={Agentsquare: Automatic llm agent search in modular design space},
  author={Shang, Yu and Li, Yu and Zhao, Keyu and Ma, Likai and Liu, Jiahe and Xu, Fengli and Li, Yong},
  journal={arXiv preprint arXiv:2410.06153},
  year={2024}
}

@article{zhang2024aflow,
  title={Aflow: Automating agentic workflow generation},
  author={Zhang, Jiayi and Xiang, Jinyu and Yu, Zhaoyang and Teng, Fengwei and Chen, Xionghui and Chen, Jiaqi and Zhuge, Mingchen and Cheng, Xin and Hong, Sirui and Wang, Jinlin and others},
  journal={arXiv preprint arXiv:2410.10762},
  year={2024}
}

@article{li2025agentswift,
  title={AgentSwift: Efficient LLM Agent Design via Value-guided Hierarchical Search},
  author={Li, Yu and Li, Lehui and Wu, Zhihao and Liao, Qingmin and Hao, Jianye and Shao, Kun and Xu, Fengli and Li, Yong},
  journal={arXiv preprint arXiv:2506.06017},
  year={2025}
}

@article{wang2025scoreflow,
  title={Scoreflow: Mastering llm agent workflows via score-based preference optimization},
  author={Wang, Yinjie and Yang, Ling and Li, Guohao and Wang, Mengdi and Aragam, Bryon},
  journal={arXiv preprint arXiv:2502.04306},
  year={2025}
}

@article{gao2025flowreasoner,
  title={Flowreasoner: Reinforcing query-level meta-agents},
  author={Gao, Hongcheng and Liu, Yue and He, Yufei and Dou, Longxu and Du, Chao and Deng, Zhijie and Hooi, Bryan and Lin, Min and Pang, Tianyu},
  journal={arXiv preprint arXiv:2504.15257},
  year={2025}
}

@article{xu2025robustflow,
  title={RobustFlow: Towards Robust Agentic Workflow Generation},
  author={Xu, Shengxiang and Zhang, Jiayi and Di, Shimin and Luo, Yuyu and Yao, Liang and Liu, Hanmo and Zhu, Jia and Liu, Fan and Zhang, Min-Ling},
  journal={arXiv preprint arXiv:2509.21834},
  year={2025}
}

@article{wang2023voyager,
  title={Voyager: An open-ended embodied agent with large language models},
  author={Wang, Guanzhi and Xie, Yuqi and Jiang, Yunfan and Mandlekar, Ajay and Xiao, Chaowei and Zhu, Yuke and Fan, Linxi and Anandkumar, Anima},
  journal={arXiv preprint arXiv:2305.16291},
  year={2023}
}

@inproceedings{qian2023creator,
  title={Creator: Tool creation for disentangling abstract and concrete reasoning of large language models},
  author={Qian, Cheng and Han, Chi and Fung, Yi and Qin, Yujia and Liu, Zhiyuan and Ji, Heng},
  booktitle={Findings of the Association for Computational Linguistics: EMNLP 2023},
  pages={6922--6939},
  year={2023}
}

@article{zheng2025skillweaver,
  title={Skillweaver: Web agents can self-improve by discovering and honing skills},
  author={Zheng, Boyuan and Fatemi, Michael Y and Jin, Xiaolong and Wang, Zora Zhiruo and Gandhi, Apurva and Song, Yueqi and Gu, Yu and Srinivasa, Jayanth and Liu, Gaowen and Neubig, Graham and others},
  journal={arXiv preprint arXiv:2504.07079},
  year={2025}
}

@article{fernando2023promptbreeder,
  title={Promptbreeder: Self-referential self-improvement via prompt evolution},
  author={Fernando, Chrisantha and Banarse, Dylan and Michalewski, Henryk and Osindero, Simon and Rockt{\"a}schel, Tim},
  journal={arXiv preprint arXiv:2309.16797},
  year={2023}
}

@article{zhang2024revolve,
  title={Revolve: Optimizing ai systems by tracking response evolution in textual optimization},
  author={Zhang, Peiyan and Jin, Haibo and Hu, Leyang and Li, Xinnuo and Kang, Liying and Luo, Man and Song, Yangqiu and Wang, Haohan},
  journal={arXiv preprint arXiv:2412.03092},
  year={2024}
}

@article{yuksekgonul2024textgrad,
  title={Textgrad: Automatic" differentiation" via text},
  author={Yuksekgonul, Mert and Bianchi, Federico and Boen, Joseph and Liu, Sheng and Huang, Zhi and Guestrin, Carlos and Zou, James},
  journal={arXiv preprint arXiv:2406.07496},
  year={2024}
}

@article{yin2025llm,
  title={LLM-AutoDiff: Auto-Differentiate Any LLM Workflow},
  author={Yin, Li and Wang, Zhangyang},
  journal={arXiv preprint arXiv:2501.16673},
  year={2025}
}

@article{khattab2023dspy,
  title={Dspy: Compiling declarative language model calls into self-improving pipelines},
  author={Khattab, Omar and Singhvi, Arnav and Maheshwari, Paridhi and Zhang, Zhiyuan and Santhanam, Keshav and Vardhamanan, Sri and Haq, Saiful and Sharma, Ashutosh and Joshi, Thomas T and Moazam, Hanna and others},
  journal={arXiv preprint arXiv:2310.03714},
  year={2023}
}

@article{chhikara2025mem0,
  title={Mem0: Building production-ready ai agents with scalable long-term memory},
  author={Chhikara, Prateek and Khant, Dev and Aryan, Saket and Singh, Taranjeet and Yadav, Deshraj},
  journal={arXiv preprint arXiv:2504.19413},
  year={2025}
}

@article{salama2025meminsight,
  title={Meminsight: Autonomous memory augmentation for llm agents},
  author={Salama, Rana and Cai, Jason and Yuan, Michelle and Currey, Anna and Sunkara, Monica and Zhang, Yi and Benajiba, Yassine},
  journal={arXiv preprint arXiv:2503.21760},
  year={2025}
}

@article{huang2024code,
  title={Da-code: Agent data science code generation benchmark for large language models},
  author={Huang, Yiming and Luo, Jianwen and Yu, Yan and Zhang, Yitong and Lei, Fangyu and Wei, Yifan and He, Shizhu and Huang, Lifu and Liu, Xiao and Zhao, Jun and others},
  journal={arXiv preprint arXiv:2410.07331},
  year={2024}
}

@inproceedings{lai2023ds,
  title={DS-1000: A natural and reliable benchmark for data science code generation},
  author={Lai, Yuhang and Li, Chengxi and Wang, Yiming and Zhang, Tianyi and Zhong, Ruiqi and Zettlemoyer, Luke and Yih, Wen-tau and Fried, Daniel and Wang, Sida and Yu, Tao},
  booktitle={International Conference on Machine Learning},
  pages={18319--18345},
  year={2023},
  organization={PMLR}
}

@article{hendrycks2021measuring,
  title={Measuring mathematical problem solving with the math dataset},
  author={Hendrycks, Dan and Burns, Collin and Kadavath, Saurav and Arora, Akul and Basart, Steven and Tang, Eric and Song, Dawn and Steinhardt, Jacob},
  journal={arXiv preprint arXiv:2103.03874},
  year={2021}
}

@article{trivedi2024appworld,
  title={Appworld: A controllable world of apps and people for benchmarking interactive coding agents},
  author={Trivedi, Harsh and Khot, Tushar and Hartmann, Mareike and Manku, Ruskin and Dong, Vinty and Li, Edward and Gupta, Shashank and Sabharwal, Ashish and Balasubramanian, Niranjan},
  journal={arXiv preprint arXiv:2407.18901},
  year={2024}
}

@article{wei2022chain,
  title={Chain-of-thought prompting elicits reasoning in large language models},
  author={Wei, Jason and Wang, Xuezhi and Schuurmans, Dale and Bosma, Maarten and Xia, Fei and Chi, Ed and Le, Quoc V and Zhou, Denny and others},
  journal={Advances in neural information processing systems},
  volume={35},
  pages={24824--24837},
  year={2022}
}

@article{zheng2023take,
  title={Take a step back: Evoking reasoning via abstraction in large language models},
  author={Zheng, Huaixiu Steven and Mishra, Swaroop and Chen, Xinyun and Cheng, Heng-Tze and Chi, Ed H and Le, Quoc V and Zhou, Denny},
  journal={arXiv preprint arXiv:2310.06117},
  year={2023}
}

@article{madaan2023self,
  title={Self-refine: Iterative refinement with self-feedback},
  author={Madaan, Aman and Tandon, Niket and Gupta, Prakhar and Hallinan, Skyler and Gao, Luyu and Wiegreffe, Sarah and Alon, Uri and Dziri, Nouha and Prabhumoye, Shrimai and Yang, Yiming and others},
  journal={Advances in Neural Information Processing Systems},
  volume={36},
  pages={46534--46594},
  year={2023}
}

@inproceedings{yang2023large,
  title={Large language models as optimizers},
  author={Yang, Chengrun and Wang, Xuezhi and Lu, Yifeng and Liu, Hanxiao and Le, Quoc V and Zhou, Denny and Chen, Xinyun},
  booktitle={The Twelfth International Conference on Learning Representations},
  year={2023}
}

@article{wang2023promptagent,
  title={Promptagent: Strategic planning with language models enables expert-level prompt optimization},
  author={Wang, Xinyuan and Li, Chenxi and Wang, Zhen and Bai, Fan and Luo, Haotian and Zhang, Jiayou and Jojic, Nebojsa and Xing, Eric P and Hu, Zhiting},
  journal={arXiv preprint arXiv:2310.16427},
  year={2023}
}

@article{yuan2023craft,
  title={Craft: Customizing llms by creating and retrieving from specialized toolsets},
  author={Yuan, Lifan and Chen, Yangyi and Wang, Xingyao and Fung, Yi R and Peng, Hao and Ji, Heng},
  journal={arXiv preprint arXiv:2309.17428},
  year={2023}
}

@article{wang2024toolgen,
  title={Toolgen: Unified tool retrieval and calling via generation},
  author={Wang, Renxi and Han, Xudong and Ji, Lei and Wang, Shu and Baldwin, Timothy and Li, Haonan},
  journal={arXiv preprint arXiv:2410.03439},
  year={2024}
}

@article{shi2025retrieval,
  title={Retrieval Models Aren't Tool-Savvy: Benchmarking Tool Retrieval for Large Language Models},
  author={Shi, Zhengliang and Wang, Yuhan and Yan, Lingyong and Ren, Pengjie and Wang, Shuaiqiang and Yin, Dawei and Ren, Zhaochun},
  journal={arXiv preprint arXiv:2503.01763},
  year={2025}
}

@article{zheng2024toolrerank,
  title={Toolrerank: Adaptive and hierarchy-aware reranking for tool retrieval},
  author={Zheng, Yuanhang and Li, Peng and Liu, Wei and Liu, Yang and Luan, Jian and Wang, Bin},
  journal={arXiv preprint arXiv:2403.06551},
  year={2024}
}

@article{schick2023toolformer,
  title={Toolformer: Language models can teach themselves to use tools},
  author={Schick, Timo and Dwivedi-Yu, Jane and Dess{\`\i}, Roberto and Raileanu, Roberta and Lomeli, Maria and Hambro, Eric and Zettlemoyer, Luke and Cancedda, Nicola and Scialom, Thomas},
  journal={Advances in Neural Information Processing Systems},
  volume={36},
  pages={68539--68551},
  year={2023}
}

@article{qin2023toolllm,
  title={Toolllm: Facilitating large language models to master 16000+ real-world apis},
  author={Qin, Yujia and Liang, Shihao and Ye, Yining and Zhu, Kunlun and Yan, Lan and Lu, Yaxi and Lin, Yankai and Cong, Xin and Tang, Xiangru and Qian, Bill and others},
  journal={arXiv preprint arXiv:2307.16789},
  year={2023}
}

@article{qiu2025alita,
  title={Alita: Generalist agent enabling scalable agentic reasoning with minimal predefinition and maximal self-evolution},
  author={Qiu, Jiahao and Qi, Xuan and Zhang, Tongcheng and Juan, Xinzhe and Guo, Jiacheng and Lu, Yifu and Wang, Yimin and Yao, Zixin and Ren, Qihan and Jiang, Xun and others},
  journal={arXiv preprint arXiv:2505.20286},
  year={2025}
}

@article{xia2025live,
  title={Live-SWE-agent: Can Software Engineering Agents Self-Evolve on the Fly?},
  author={Xia, Chunqiu Steven and Wang, Zhe and Yang, Yan and Wei, Yuxiang and Zhang, Lingming},
  journal={arXiv preprint arXiv:2511.13646},
  year={2025}
}

@inproceedings{zhao2024expel,
  title={Expel: Llm agents are experiential learners},
  author={Zhao, Andrew and Huang, Daniel and Xu, Quentin and Lin, Matthieu and Liu, Yong-Jin and Huang, Gao},
  booktitle={Proceedings of the AAAI Conference on Artificial Intelligence},
  volume={38},
  number={17},
  pages={19632--19642},
  year={2024}
}

@article{liang2024self,
  title={Self-evolving Agents with reflective and memory-augmented abilities},
  author={Liang, Xuechen and He, Yangfan and Xia, Yinghui and Song, Xinyuan and Wang, Jianhui and Tao, Meiling and Sun, Li and Yuan, Xinhang and Su, Jiayi and Li, Keqin and others},
  journal={arXiv preprint arXiv:2409.00872},
  year={2024}
}

@article{wang2024agent,
  title={Agent workflow memory},
  author={Wang, Zora Zhiruo and Mao, Jiayuan and Fried, Daniel and Neubig, Graham},
  journal={arXiv preprint arXiv:2409.07429},
  year={2024}
}

@article{pei2025scope,
  title={SCOPE: Prompt Evolution for Enhancing Agent Effectiveness},
  author={Pei, Zehua and Zhen, Hui-Ling and Kai, Shixiong and Pan, Sinno Jialin and Wang, Yunhe and Yuan, Mingxuan and Yu, Bei},
  journal={arXiv preprint arXiv:2512.15374},
  year={2025}
}

@article{haque2025advanced,
  title={Advanced Tool Learning and Selection System (ATLASS): A Closed-Loop Framework Using LLM},
  author={Haque, Mohd Ariful and Williams, Justin and Siddique, Sunzida and Islam, Md Hujaifa and Ali, Hasmot and Gupta, Kishor Datta and George, Roy},
  journal={arXiv preprint arXiv:2503.10071},
  year={2025}
}

@inproceedings{min2022metaicl,
  title={Metaicl: Learning to learn in context},
  author={Min, Sewon and Lewis, Mike and Zettlemoyer, Luke and Hajishirzi, Hannaneh},
  booktitle={Proceedings of the 2022 conference of the North American chapter of the Association for Computational Linguistics: Human Language Technologies},
  pages={2791--2809},
  year={2022}
}

@inproceedings{zhou2022large,
  title={Large language models are human-level prompt engineers},
  author={Zhou, Yongchao and Muresanu, Andrei Ioan and Han, Ziwen and Paster, Keiran and Pitis, Silviu and Chan, Harris and Ba, Jimmy},
  booktitle={The eleventh international conference on learning representations},
  year={2022}
}

@inproceedings{yi2025zera,
  title={ZERA: Zero-init Instruction Evolving Refinement Agent--From Zero Instructions to Structured Prompts via Principle-based Optimization},
  author={Yi, Seungyoun and Khang, Minsoo and Park, Sungrae},
  booktitle={Proceedings of the 2025 Conference on Empirical Methods in Natural Language Processing},
  pages={23334--23348},
  year={2025}
}

@inproceedings{meirelesinfluence,
  title={The Influence of Scaffolds on Coordination Scaling Laws in LLM Agents},
  author={Meireles, Mariana and Bhati, Rupali and Lauffer, Niklas and Allen, Cameron},
  booktitle={Workshop on Scaling Environments for Agents},
  year={2025}
}

@article{chollet2019measure,
  title={On the measure of intelligence},
  author={Chollet, Fran{\c{c}}ois},
  journal={arXiv preprint arXiv:1911.01547},
  year={2019}
}

@article{li2024numinamath,
  title={Numinamath: The largest public dataset in ai4maths with 860k pairs of competition math problems and solutions},
  author={Li, Jia and Beeching, Edward and Tunstall, Lewis and Lipkin, Ben and Soletskyi, Roman and Huang, Shengyi and Rasul, Kashif and Yu, Longhui and Jiang, Albert Q and Shen, Ziju and others},
  journal={Hugging Face repository},
  volume={13},
  number={9},
  pages={9},
  year={2024}
}

@article{zhang2026expseek,
  title={ExpSeek: Self-Triggered Experience Seeking for Web Agents},
  author={Zhang, Wenyuan and Zhang, Xinghua and Yu, Haiyang and Nie, Shuaiyi and Wu, Bingli and Yue, Juwei and Liu, Tingwen and Li, Yongbin},
  journal={arXiv preprint arXiv:2601.08605},
  year={2026}
}

@inproceedings{ma2026talk2image,
  title={Talk2image: A multi-agent system for multi-turn image generation and editing},
  author={Ma, Shichao and Guo, Yunhe and Su, Jiahao and Huang, Qihe and Zhou, Zhengyang and Wang, Yang},
  booktitle={Proceedings of the AAAI Conference on Artificial Intelligence},
  volume={40},
  number={38},
  pages={32437--32445},
  year={2026}
}

@article{liu2025spark,
  title={SPARK: Synergistic Policy And Reward Co-Evolving Framework},
  author={Liu, Ziyu and Zang, Yuhang and Ding, Shengyuan and Cao, Yuhang and Dong, Xiaoyi and Duan, Haodong and Lin, Dahua and Wang, Jiaqi},
  journal={arXiv preprint arXiv:2509.22624},
  year={2025}
}

@misc{li2026whatsmissingscreentoactionuiintheloop,
      title={What's Missing in Screen-to-Action? Towards a UI-in-the-Loop Paradigm for Multimodal GUI Reasoning}, 
      author={Songze Li and Xiaoke Guo and Tianqi Liu and Biao Yi and Zhaoyan Gong and Zhiqiang Liu and Huajun Chen and Wen Zhang},
      year={2026},
      eprint={2604.06995},
      archivePrefix={arXiv},
      primaryClass={cs.AI},
      url={https://arxiv.org/abs/2604.06995}, 
}

@misc{zhang2026dontactblindlyrobust,
      title={Don't Act Blindly: Robust GUI Automation via Action-Effect Verification and Self-Correction}, 
      author={Yuzhe Zhang and Xianwei Xue and Xingyong Wu and Mengke Chen and Chen Liu and Xinran He and Run Shao and Feiran Liu and Huanmin Xu and Qiutong Pan and Haiwei Wang},
      year={2026},
      eprint={2604.05477},
      archivePrefix={arXiv},
      primaryClass={cs.CL},
      url={https://arxiv.org/abs/2604.05477}, 
}

@misc{long2026emomasemotionawaremultiagenthighstakes,
      title={EmoMAS: Emotion-Aware Multi-Agent System for High-Stakes Edge-Deployable Negotiation with Bayesian Orchestration}, 
      author={Yunbo Long and Yuhan Liu and Liming Xu},
      year={2026},
      eprint={2604.07003},
      archivePrefix={arXiv},
      primaryClass={cs.AI},
      url={https://arxiv.org/abs/2604.07003}, 
}

@misc{bai2026ttvsboostingselfexploringreinforcement,
      title={TTVS: Boosting Self-Exploring Reinforcement Learning via Test-time Variational Synthesis}, 
      author={Sikai Bai and Haoxi Li and Jie Zhang and Yongjiang Liu and Song Guo},
      year={2026},
      eprint={2604.08468},
      archivePrefix={arXiv},
      primaryClass={cs.LG},
      url={https://arxiv.org/abs/2604.08468}, 
}

@article{lin2025se,
  title={Se-agent: Self-evolution trajectory optimization in multi-step reasoning with llm-based agents},
  author={Lin, Jiaye and Guo, Yifu and Han, Yuzhen and Hu, Sen and Ni, Ziyi and Wang, Licheng and Chen, Mingguang and Liu, Hongzhang and Chen, Ronghao and He, Yangfan and others},
  journal={arXiv preprint arXiv:2508.02085},
  year={2025}
}

@misc{zhang2026couplingmacrodynamicsmicro,
      title={Coupling Macro Dynamics and Micro States for Long-Horizon Social Simulation}, 
      author={Yunyao Zhang and Yihao Ai and Zuocheng Ying and Qirui Mi and Junqing Yu and Wei Yang and Zikai Song},
      year={2026},
      eprint={2604.05516},
      archivePrefix={arXiv},
      primaryClass={cs.SI},
      url={https://arxiv.org/abs/2604.05516}, 
}

@inproceedings{zhang-etal-2025-ga,
    title = "$GA-S^3$: Comprehensive Social Network Simulation with Group Agents",
    author = "Zhang, Yunyao  and
      Song, Zikai  and
      Zhou, Hang  and
      Ren, Wenfeng  and
      Chen, Yi-Ping Phoebe  and
      Yu, Junqing  and
      Yang, Wei",
    editor = "Che, Wanxiang  and
      Nabende, Joyce  and
      Shutova, Ekaterina  and
      Pilehvar, Mohammad Taher",
    booktitle = "Findings of the Association for Computational Linguistics: ACL 2025",
    month = jul,
    year = "2025",
    address = "Vienna, Austria",
    publisher = "Association for Computational Linguistics",
    url = "https://aclanthology.org/2025.findings-acl.468/",
    doi = "10.18653/v1/2025.findings-acl.468",
    pages = "8950--8970",
    ISBN = "979-8-89176-256-5"
}

@article{ouyang2024learn,
  title={Learn from global correlations: Enhancing evolutionary algorithm via spectral gnn},
  author={Ouyang, Kaichen and Ke, Zong and Fu, Shengwei and Liu, Lingjie and Zhao, Puning and Hu, Dayu},
  journal={arXiv preprint arXiv:2412.17629},
  year={2024}
}

@article{ke2025early,
  title={Early warning of cryptocurrency reversal risks via multi-source data},
  author={Ke, Zong and Cao, Yuqing and Chen, Zhenrui and Yin, Yuchen and He, Shouchao and Cheng, Yu},
  journal={Finance Research Letters},
  pages={107890},
  year={2025},
  publisher={Elsevier}
}

@article{chen2023bias,
  title={Bias and debias in recommender system: A survey and future directions},
  author={Chen, Jiawei and Dong, Hande and Wang, Xiang and Feng, Fuli and Wang, Meng and He, Xiangnan},
  journal={ACM Transactions on Information Systems},
  volume={41},
  number={3},
  pages={1--39},
  year={2023},
  publisher={ACM New York, NY}
}

@inproceedings{lin2025recommendation,
  title={How do recommendation models amplify popularity bias? An analysis from the spectral perspective},
  author={Lin, Siyi and Gao, Chongming and Chen, Jiawei and Zhou, Sheng and Hu, Binbin and Feng, Yan and Chen, Chun and Wang, Can},
  booktitle={Proceedings of the Eighteenth ACM International Conference on Web Search and Data Mining},
  pages={659--668},
  year={2025}
}

@article{cui2025hatllm,
  title={HatLLM: Hierarchical Attention Masking for Enhanced Collaborative Modeling in LLM-based Recommendation},
  author={Cui, Yu and Liu, Feng and Chen, Jiawei and Jin, Canghong and Lou, Xingyu and Zhang, Changwang and Wang, Jun and Sun, Yuegang and Wang, Can},
  journal={arXiv preprint arXiv:2510.10955},
  year={2025}
}

@article{gao2023cirs,
  title={CIRS: Bursting filter bubbles by counterfactual interactive recommender system},
  author={Gao, Chongming and Wang, Shiqi and Li, Shijun and Chen, Jiawei and He, Xiangnan and Lei, Wenqiang and Li, Biao and Zhang, Yuan and Jiang, Peng},
  journal={ACM Transactions on Information Systems},
  volume={42},
  number={1},
  pages={1--27},
  year={2023},
  publisher={ACM New York, NY, USA}
}

@inproceedings{gao2023alleviating,
  title={Alleviating matthew effect of offline reinforcement learning in interactive recommendation},
  author={Gao, Chongming and Huang, Kexin and Chen, Jiawei and Zhang, Yuan and Li, Biao and Jiang, Peng and Wang, Shiqi and Zhang, Zhong and He, Xiangnan},
  booktitle={Proceedings of the 46th international ACM SIGIR conference on research and development in information retrieval},
  pages={238--248},
  year={2023}
}

@misc{cheng2026mem2evolveselfevolvingagentscoevolutionary,
      title={Mem$^2$Evolve: Towards Self-Evolving Agents via Co-Evolutionary Capability Expansion and Experience Distillation}, 
      author={Zihao Cheng and Zeming Liu and Yingyu Shan and Xinyi Wang and Xiangrong Zhu and Yunpu Ma and Hongru Wang and Yuhang Guo and Wei Lin and Yunhong Wang},
      year={2026},
      eprint={2604.10923},
      archivePrefix={arXiv},
      primaryClass={cs.CL},
      url={https://arxiv.org/abs/2604.10923}, 
}

@article{chang2024survey,
  title={A survey on evaluation of large language models},
  author={Chang, Yupeng and Wang, Xu and Wang, Jindong and Wu, Yuan and Yang, Linyi and Zhu, Kaijie and Chen, Hao and Yi, Xiaoyuan and Wang, Cunxiang and Wang, Yidong and others},
  journal={ACM transactions on intelligent systems and technology},
  volume={15},
  number={3},
  pages={1--45},
  year={2024},
  publisher={ACM New York, NY}
}

\clearpage
\appendix

\appendixtableofcontents

\clearpage

\appsection{Usage of LLMs}

Throughout the preparation of this manuscript, Large Language Models (LLMs) were utilized as a writing and editing tool. Specifically, we employed LLMs to improve the clarity and readability of the text, refine sentence structures, and correct grammatical errors. All final content, including the core scientific claims, experimental design, and conclusions, was conceived and written by us, and we take full responsibility for the final version of this paper.

\appsection{Related Work}

\appsubsection{Automated Agentic Generation}

Automated agent generation methods can be roughly divided into two lines: they either search an agent from a predefined module pool or train a scaffold generator.
ADAS~\citep{hu2024automated} and AFlow~\citep{zhang2024aflow} treat an agent as a program or workflow and use a meta-agent or MCTS-style search to iteratively propose, execute, and retain high-scoring designs in a hand-crafted search space.
AgentSquare~\citep{shang2024agentsquare} abstracts agents into four interchangeable modules (planning, reasoning, tool use, memory), while AgentSwift~\citep{li2025agentswift} further enlarges the space by jointly searching workflow structure and functional components under a value-guided, uncertainty-aware hierarchical search.
These search-based methods operate over increasingly rich design spaces but still rely on coarse scalar scores, without explicitly reasoning over the interaction experience when updating scaffold.

Another line of approaches learn an LLM policy that generates scaffolds.
ScoreFlow~\citep{wang2025scoreflow} trains a workflow generator with a score-based preference objective, turning workflow optimization into learning from pairwise preferences induced by evaluation scores.
RobustFlow~\citep{xu2025robustflow} extends this view to robustness, optimizing generators so that workflows remain consistent across perturbed but semantically equivalent instructions.
FlowReasoner~\citep{gao2025flowreasoner} instead optimizes a query-level meta-agent with external execution feedback, using distillation plus reinforcement learning to improve the multi-agent systems it designs for each query.

ReCreate differs from both lines by taking full interaction experience (trajectories, logs, execution artifacts, verifier outputs) as input to a ReCreate-Agent that proposes targeted scaffold edits and enables experience-grounded agent optimization.

\appsubsection{Self-Evolve Methods}

\paragraph{Automated Tool Learning}
A prominent line of self-evolving agents enhances what an agent can do by autonomously expanding and maintaining its tool set.
In embodied scenarios, long-horizon settings, Voyager~\citep{wang2023voyager} continually explores and accumulates reusable skills, forming a growing library of executable behaviors. For more general-purpose tool creation, works such as Alita~\citep{qiu2025alita}, Live-SWE-Agent~\citep{xia2025live}, ATLASS~\citep{haque2025advanced}, CREATOR~\citep{qian2023creator}, SkillWeaver~\citep{zheng2025skillweaver}, and CRAFT~\citep{yuan2023craft} generate new functions or APIs from interaction experience and execution feedback, and reuse them across tasks. Beyond tool creation, an additional challenge is tool selection under a large inventory: methods such as ToolGen~\citep{wang2024toolgen}, ToolRet~\citep{shi2025retrieval}, and ToolRerank~\citep{zheng2024toolrerank} retrieve, rerank, and invoke appropriate tools more reliably. Tool learning is also studied at the level of tool-use competence, e.g., Toolformer~\citep{schick2023toolformer} and ToolLLM~\citep{qin2023toolllm}, which train or distill tool-use behaviors to improve tool calling accuracy and robustness.

\paragraph{Automated Context Learning}
Another core direction evolves what an agent sees in its context window, most commonly through prompt and instruction optimization. Early representative approaches treat prompt search as a discrete optimization problem: APE~\citep{zhou2022large} and MetaICL~\citep{min2022metaicl} generate candidate prompts and select among them based on validation performance. More agentic variants explicitly plan over the prompt space, such as PromptAgent~\citep{wang2023promptagent}, while population-based evolution is exemplified by PromptBreeder~\citep{fernando2023promptbreeder}. 
To stabilize and accelerate iterative prompt refinement, OPRO~\citep{yang2023large} and REVOLVE~\citep{zhang2024revolve} use model-generated critiques and edits as optimization steps; similarly, ZERA~\citep{yi2025zera} performs training-free evaluation--refinement with principle-based critiques and jointly refines system and user prompts (and task descriptions). For agentic settings with long and dynamic traces, SCOPE~\citep{pei2025scope} treats prompt evolution as an online optimization problem and updates prompts from execution traces. Beyond single prompts, pipeline-level context learning is captured by DSPy~\citep{khattab2023dspy}, and gradient-style textual optimization is explored in TextGrad~\citep{yuksekgonul2024textgrad}, LLM-AutoDiff~\citep{yin2025llm} and others~\citep{liu2025spark, chen2023bias, lin2025recommendation, cui2025hatllm, gao2023cirs, gao2023alleviating}. Overall, these methods optimize the in-context specification (instructions, exemplars, and intermediate prompts) to steer agent behavior, and are largely orthogonal to expanding the toolset or updating long-term memory.

\paragraph{Automated Memory Evolving}
Memory evolving methods update what an agent retains and retrieves across episodes by deciding what to store, revise, and discard. One line focuses on structured long-term memory maintenance: SAGE~\citep{liang2024self} uses a forgetting-curve-inspired retention heuristic, while Mem0~\citep{chhikara2025mem0} and MemInsight~\citep{salama2025meminsight} use explicit update operations and semantic organization to support retrieval. Another line treats memory as an experience library by summarizing interaction history into reusable guidance: Expel~\citep{zhao2024expel} distills trajectories into actionable rules, and Agent Workflow Memory~\citep{wang2024agent} stores workflow fragments that can be replayed for similar tasks. Memory evolution is also explored in multi-agent interaction settings: self-play accumulates negotiation knowledge over time~\citep{cheng2026mem2evolveselfevolvingagentscoevolutionary, long2026emomasemotionawaremultiagenthighstakes, lin2025se,ouyang2024learn,ke2025early}, while social simulation frameworks leverage persistent group memory to model long-horizon agent dynamics~\citep{zhang2026couplingmacrodynamicsmicro, zhang-etal-2025-ga, ma2026talk2image, zhang2026expseek, bai2026ttvsboostingselfexploringreinforcement}. Overall, these approaches treat memory as a persistent object that is continually updated and consulted to guide future decisions.


\appsection{Discussions}
\label{sec:discussion}

\paragraph{Why Agent-as-optimizer?}
While the concept of \emph{LLM-as-optimizer} is widely recognized~\citep{yuksekgonul2024textgrad}, \emph{Agent-as-optimizer} remains an emerging frontier.
We identify \emph{domain agent creation} as a quintessential scenario to exemplify this distinction.
Fundamentally, \emph{Agent-as-optimizer}  represents a paradigm shift from \emph{Optimization by Prompting} to \emph{Optimization by Doing}.
The former follows a linear \texttt{Reasoning $\to$ Text} process, passively generating prompts based on static context.
Crucially, this approach remains labor-intensive, as it requires humans to manually curate and feed specific optimization targets into the model's context.
In contrast, ReCreate establishes an {Optimization by Doing} loop: \texttt{Inspect $\to$ Reason $\to$ Optimize}.
Here, ReCreate-Agent acts as an autonomous engineer: it actively retrieves specific trajectories, execution diffs or evaluation results to diagnose failure modes.
This shifts the paradigm from reading static history to navigating full experience, enabling the precise localization of bug roots hidden in massive logs.

\paragraph{Comparison to Self-Evolve}
Recent self-evolving methods~\citep{xia2025live, yang2023large, zhao2024expel} also leverage experience to refine pre-existing agents. ReCreate differs from these self-evolving methods in three aspects:
(1) \emph{Scope:} Instead of iteratively polishing a pre-defined scaffold, ReCreate can \emph{create} an agent from scratch, which makes it applicable even in domains where no mature agent or hand-crafted workflow is available. 
(2) \emph{Objective:} While prior work mainly optimizes for instance-level success (i.e., improving performance on the specific tasks encountered), ReCreate targets \emph{domain-level generalization} by abstracting reusable improvements through hierarchical updates, thereby reducing overfitting to individual instances.
(3) \emph{Strategy:} Rather than relying primarily on coarse outcome signals (e.g., pass/fail or scalar rewards), ReCreate performs \emph{fine-grained inspection} of execution traces and environment feedback, and turns concrete evidence into grounded scaffold edits.
Empirically, even when initialized with a minimal seed scaffold, ReCreate outperforms these methods that start from fully-developed scaffolds (\cf Section~5).

\appsection{Implementations of ReCreate}
\label{sec:implementations}
We implement \textsc{ReCreate} as a parallel evolution pipeline that improves a shared scaffold (prompt + tools + memory) across iterations.
For each batch, a task agent runs multiple instances in parallel under the same scaffold inside Docker, and we record the trajectory, submitted artifacts (e.g., patches/files), and the evaluator report.
A per-instance meta-agent then inspects these artifacts and produces a concrete update (a scaffold diff, a new tool, or a memory entry), and a synthesis meta-agent merges updates from the whole batch into the next scaffold version while removing instance-specific changes.
To support five benchmarks with one codebase, we use a \texttt{DomainAdapter} that only specifies how to load data, run the agent, and evaluate, while the evolution logic stays identical across domains.
The entire process is logged as versioned folders (\texttt{global\_v000}, \texttt{global\_v001}, \dots) with diffs and statistics, enabling reproducible comparisons to the baseline scaffold.

Following \emph{Agent Skills} design~\citep{skills}, we package each tool as a self-contained directory with a \texttt{SKILL.md} (YAML \texttt{name}/\texttt{description} for discovery) plus executable scripts and optional resources.
The agent only preloads lightweight metadata, and lazily reads the full instructions or runs scripts on demand, enabling many tools without saturating the context window.
In our system, \textsc{ReCreate}-Agent creates and updates these skill-style tool folders from execution evidence (trajectories, artifacts, and evaluator logs), so improvements are reusable and traceable to concrete failures or successful patterns.

As for memory, we implement two complementary components: a \emph{memory mechanism} and a \emph{static memory bank}.
The memory mechanism specifies when the task agent should write new memories and when it should retrieve  existing memories (e.g., after repeated failures or before critical steps), making memory usage a controlled part of the workflow.
The static memory bank stores reusable experience distilled by \textsc{ReCreate}-Agent (e.g., common failure modes, repair heuristics, and tool-usage tips), which can be searched and reused across future instances.

\begin{figure*}[htbp]
    \centering
    \includegraphics[width=0.6\linewidth, height=4.8cm]{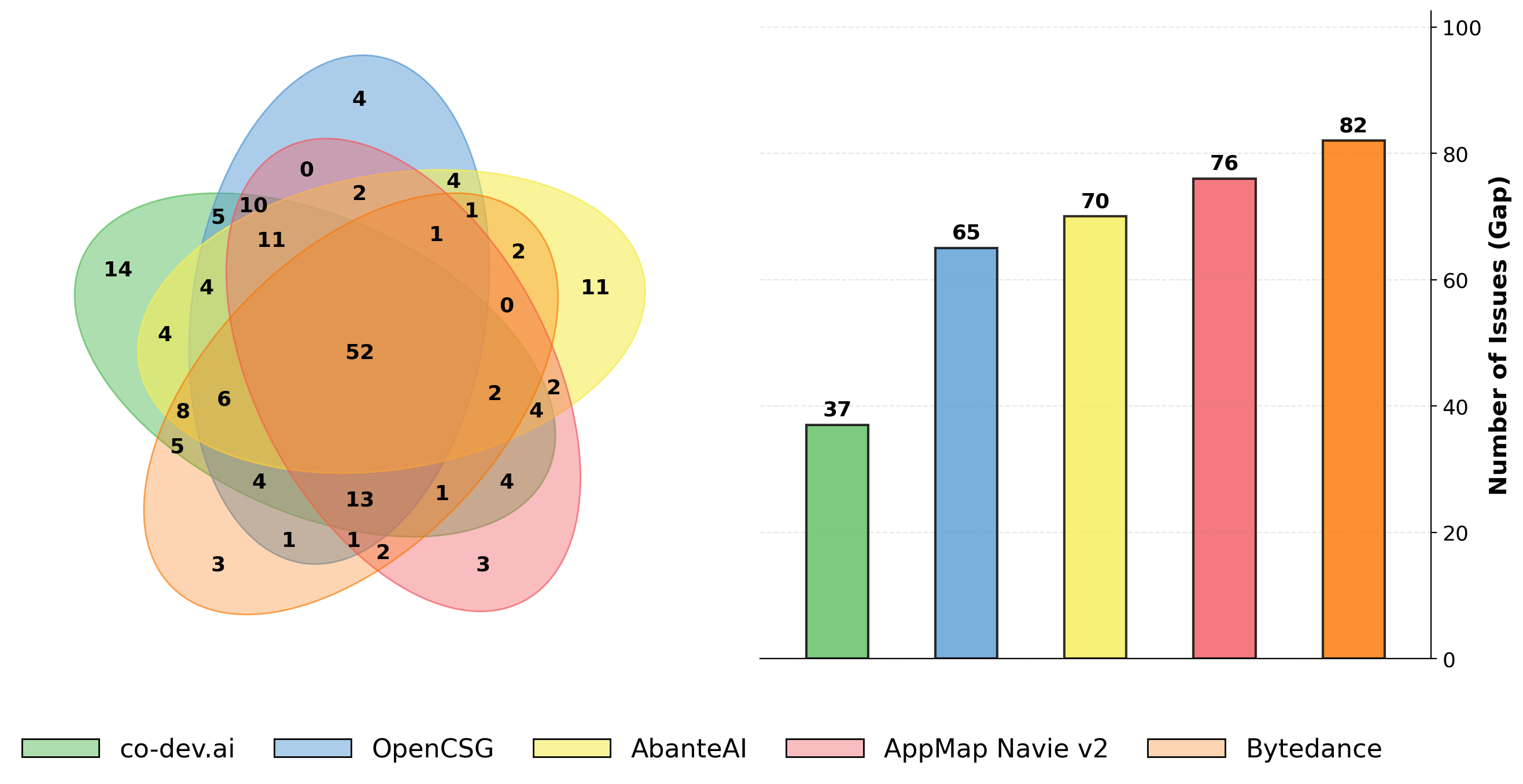}
    \caption{Scaffolds gate what a base model can do.}
    \label{Venn diagram}
\end{figure*}

\appsection{Limitations of Human-designed Scaffolds}
\label{sec:human_scaffold_limits}
In this section, we argue that human-designed scaffolds are not only labor-intensive to build, but also \emph{cap performance}. 

\paragraph{Current agent scaffolds gate what a base model can do.}
We ask a simple question: \emph{for a fixed base model, how much can the final success depend on the surrounding scaffold?}
To isolate the effect of scaffolds, we compare five top-performing open-sourced agents on SWE-bench Lite (300 issues) that all use the same LLM (gpt-4o) but differ in prompts, workflows, and tool setups.
Figure~\ref{Venn diagram} (left) shows that their solved issues overlap only partially: the union reaches 184 issues, yet the best single scaffold solves 147.
This leaves a \emph{scaffold-fixable headroom} of $184-147=37$ issues (20\% of the union): these issues are solved by the \emph{same} model under some scaffold, but are missed by the strongest human-designed scaffold in this pool.
The small intersection (only 52 issues solved by all five) further suggests that scaffolds do not merely guide outputs: they also change the agent’s search behavior (what to inspect, which checks to run, how to iterate), effectively routing the model to different solvable regions.

The right panel quantifies this effect by counting, for each scaffold $S$, how many issues in the union $U$ it fails to solve (i.e., issues solved by at least one other scaffold).
Even for the best scaffold, 37 union issues are missed; for other scaffolds, the gap is much larger (65--82 issues).
In other words, a substantial portion of what looks like “model limitation” under one scaffold is actually \emph{recoverable} under another scaffold with the same base model.
This exposes a key weakness of human-designed scaffolds: they are strong but incomplete samples from a vast design space, and they leave significant performance untapped.
These observations motivate us to perform agent scaffold optimization rather than one-off manual engineering.
Moreover, as AI continues to surpass human intelligence, it can achieve a higher upper bound in creating agents.


\begin{table*}[htbp]
\centering
\small
\setlength{\tabcolsep}{6pt}
\renewcommand{\arraystretch}{1.25}
\resizebox{\textwidth}{!}{%
\begin{tabular}{|c|c|c|c|c|c|c|c|}
\hline
\multicolumn{2}{|c|}{\textbf{SWE}} &
\multicolumn{3}{c|}{\textbf{DA-Code}} &
\multicolumn{3}{c|}{\textbf{DS-1000}} \\
\hline
\textbf{Django} & \textbf{Sympy} &
\textbf{Data Wrangling} & \textbf{Machine Learning} & \textbf{Statistical Analysis} &
\textbf{NumPy} & \textbf{Pandas} & \textbf{Matplotlib} \\
\hline
231 & 75 & 100 & 100 & 78 & 220 & 291 & 155 \\
\hline
\multicolumn{5}{|c|}{\textbf{Math}} &
\multicolumn{3}{c|}{\textbf{AppWorld}} \\
\hline
\textbf{AIME24} & \textbf{AIME25} & \textbf{Algebra} & \textbf{Number Theory} & \textbf{Counting\&Probability} &
\textbf{dev} & \textbf{Normal} & \textbf{Challenge} \\
\hline
30 & 30 & 124 & 62 & 38 & 57 & 168 & 417 \\
\hline
\end{tabular}%
}
\caption{The counts for datasets used in our experiments.}
\label{The counts for dataset used in our experiments.}
\end{table*}

\appsection{Detailed Information for Datasets}\label{Detailed Information for Datasets}
Our experiments are conducted on five benchmarks (counts in Table~\ref{The counts for dataset used in our experiments.}); we briefly introduce each benchmark below.

\noindent\textbf{SWE}~\citep{jimenez2023swe} We use \textsc{SWE-bench Verified}, where each instance corresponds to a real GitHub issue in a target repository.
The agent must produce a code patch that is applied and validated in an isolated environment; success is determined by passing the benchmark’s tests after patch application.
Django and SymPy are the two repositories in SWE-bench Verified that cover the largest number of tasks.
In our experiments, we sample 20 tasks from each repository as the development set.

\noindent\textbf{DA-Code}~\citep{huang2024code} This benchmark targets data-science programming workflows, covering common routines such as data transformation/cleaning, classical ML modeling, and statistical analysis. Tasks emphasize producing executable code under practical workflow constraints.
In our experiments, we sample 20 tasks from each subset as the development set.

\noindent\textbf{DS-1000}~\citep{lai2023ds} \textsc{DS-1000} is a data-science code generation benchmark built from real-world questions, paired with automatic evaluation via executable checks. We report results on three core library subsets that represent array computing (NumPy), tabular manipulation (Pandas), and visualization (Matplotlib). 
In our experiments, we sample 20 tasks from each repository as the development set.

\noindent\textbf{Math} AIME24  and AIME25  ~\citep{li2024numinamath} contain problems in AIME exams of the corresponding years and evaluate competitive-math reasoning with short final answers. 
We additionally use MATH500 ~\citep{hendrycks2021measuring}, a 500-problem subset of the MATH dataset, and break it down into topic subsets (Algebra, Number Theory, and Counting \& Probability) to study subject-specific behavior. 
In our experiments, we use AIME24 as the development set and evaluate on the MATH500 topic subsets as test sets.

\noindent\textbf{AppWorld}~\citep{trivedi2024appworld} \textsc{AppWorld} is an interactive agent benchmark with a suite of apps and executable APIs. Tasks require multi-step decision making and tool use in a controlled environment. We follow its provided split into a dev set and two evaluation partitions (Normal and Challenge), where the latter typically poses harder or more adversarial scenarios. In our experiments, we sample 30 instances from the dev split as the development set, and evaluate on the Normal and Challenge splits as test sets.

\appsection{Temperature Sensitivity}
We test the performance of ReCreate with different sampling temperature $t$ of ReCreate Agent (we use \texttt{claude-4.5-opus}), as shown in Table~\ref{ReCreate performance with different temperature.}.
It can be observed that ReCreate maintains comparable performance across different sampling temperatures. This suggests that state-of-the-art models have stably approached the capability to create domain agents.

\begin{table}[t]
\centering
\setlength{\tabcolsep}{5pt}
\renewcommand{\arraystretch}{1.1}
{\fontsize{9}{11}\selectfont
\begin{tabular}{ll|ccc}
\toprule
\textbf{Domain} & \textbf{Dataset}  &$t=0.0$    &$t=0.5$    &$t=1.0$\\
\toprule
\multirow{2}{*}{SWE}
& \emph{Django}      & 60.66 & 61.14 & 63.51 \\
& \emph{Sympy}       & 61.82 & 58.18 & 60.00 \\
\hline
\multirow{6}{*}{DS}
& \emph{DW}          & 51.94 & 47.79 & 50.55 \\
& \emph{ML}          & 40.88 & 51.16 & 45.78 \\
& \emph{SA}          & 23.00 & 20.33 & 27.33 \\
& \emph{Numpy}       & 77.00 & 78.00 & 81.50 \\
& \emph{Pandas}      & 67.53 & 65.68 & 66.79 \\
& \emph{Matplotlib}  & 85.19 & 78.52 & 74.07 \\
\hline
\multirow{3}{*}{Math}
& \emph{Algebra}     & 95.16 & 94.35 & 93.55 \\
& \emph{NT}          & 100.00 & 100.00 & 100.00 \\
& \emph{C\&P}        & 100.00 & 100.00 & 97.37 \\
\hline
\multirow{2}{*}{Digital}
& \emph{Normal}      & 52.98 & 51.79 & 55.36 \\
& \emph{Challenge}   & 37.41 & 37.65 & 38.85 \\
\hline
\multicolumn{2}{c|}{\textbf{Average}}  & 65.66 & 64.97 & 65.74 \\
\bottomrule
\end{tabular}
}
\caption{ReCreate performance with different temperature.}
\label{ReCreate performance with different temperature.}
\end{table}

\begin{figure*}[t]
    \centering
    \includegraphics[width=0.99\linewidth]{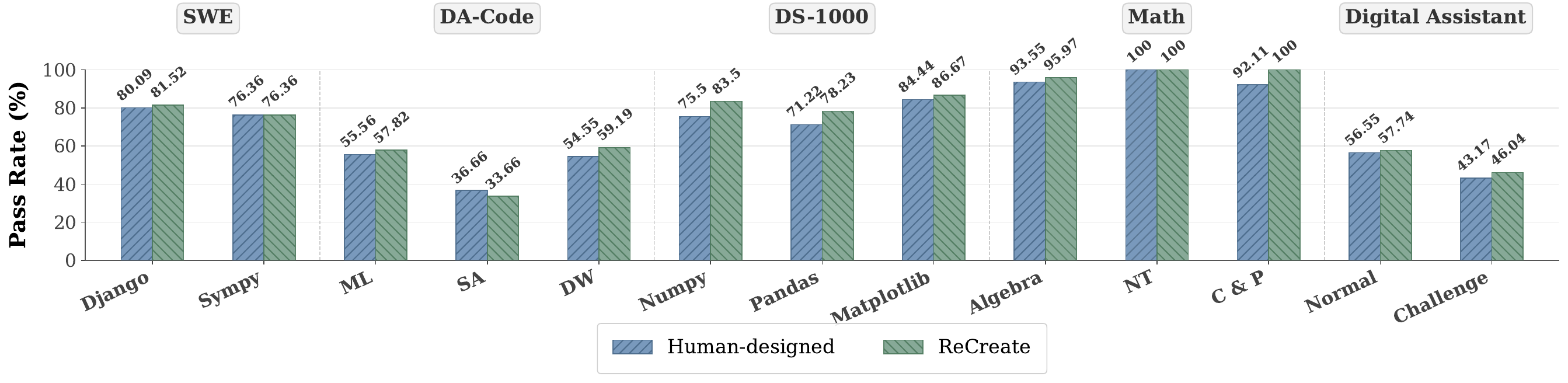}
    \caption{ReCreate on various base models for ReCreate Agent.}
    \label{ReCreate on various base models for ReCreate Agent.}
\end{figure*}

\begin{figure*}[t]
    \centering
    \includegraphics[width=0.98\linewidth]{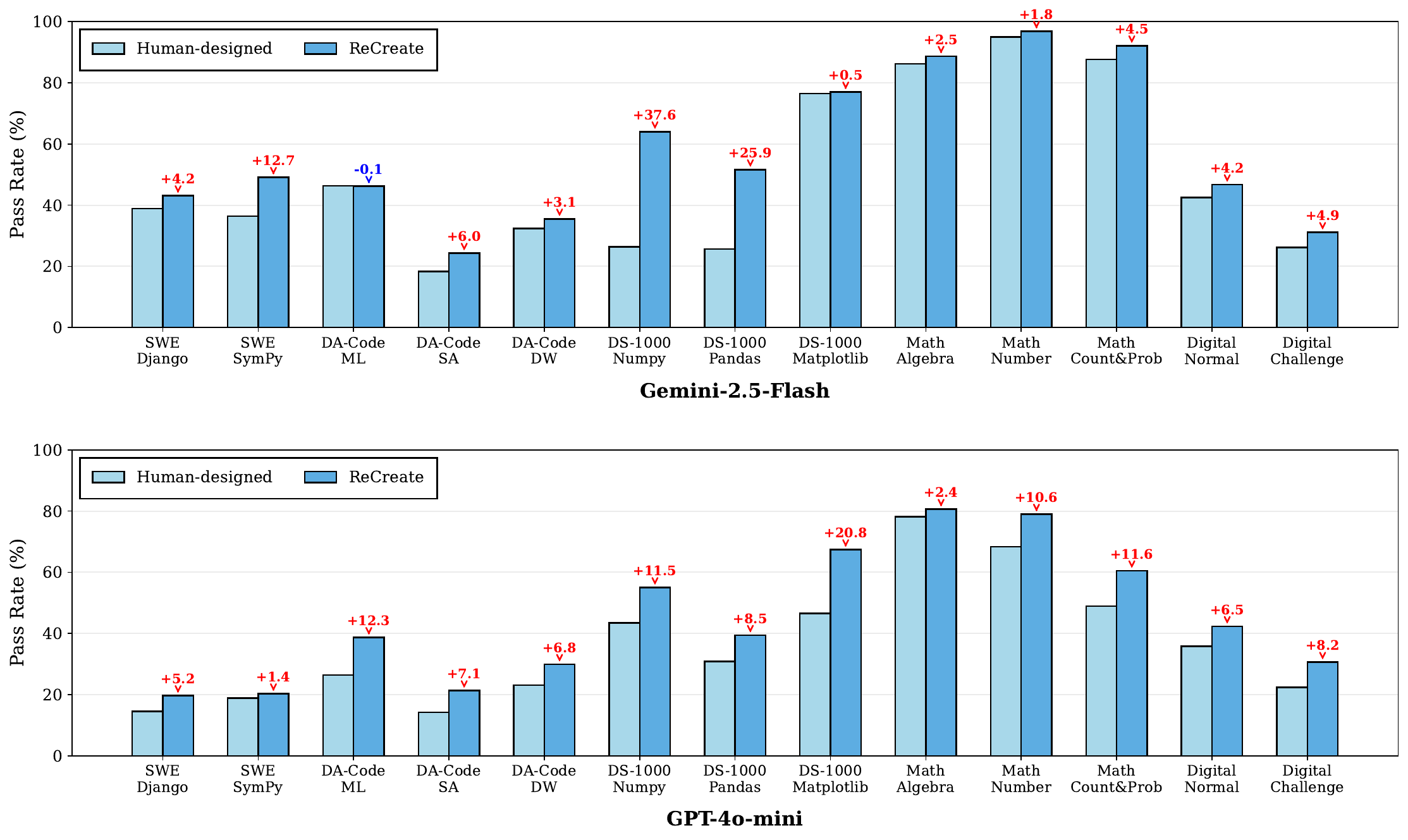}
    \caption{ReCreate on various base models for task agent.}
    \label{ReCreate on various base models for task agent.}
\end{figure*}


\appsection{Generalization on Various Models}
\paragraph{Generalization on Various Base Models for ReCreate Agent}
One may suspect that ReCreate’s gains in the main results simply stem from the stronger reasoning capability of the selected ReCreate-Agent.
We additionally evaluate ReCreate with Claude Opus 4.5 as both the ReCreate-Agent (Meta-Agent) and the task agent. The results are shown in Figure~\ref{ReCreate on various base models for ReCreate Agent.}. ReCreate remains consistently better than baseline across all evaluated domains.
As can be seen, ReCreate consistently outperforms the human-designed agent across the four domains, demonstrating that our experience-driven, automatically designed paradigm can effectively replace carefully hand-crafted designs in many settings.

\paragraph{Generalization on Various Base Models for Task Agent}
Beyond the default backbone (GPT-5-mini) in our main experiments, we further evaluate ReCreate by swapping the base LLM from GPT-5-mini to Gemini-2.5-Flash and GPT-4o-mini.
As shown in Figure~\ref{ReCreate on various base models for task agent.}, ReCreate consistently outperforms the human-designed scaffold (12/13 tasks on Gemini and 13/13 on GPT-4o-mini), delivering +8.3 and +8.7 absolute pass-rate points on average, with the largest gains on DS-1000 (up to +37.6). These results indicate that ReCreate’s experience-driven scaffold edits capture largely model-agnostic role/process/tool/memory patterns rather than backbone-specific prompt tricks.

\appsection{Significance Test}
While some test sets in the main results are relatively small (e.g., 38--75 instances), many datasets are substantially larger: for example, Pandas, Numpy, and AppWorld Challenge contain 271, 200, and 417 instances, respectively.

To assess whether ReCreate's improvements are statistically significant, we conduct McNemar paired significance tests on all test instances across 13 datasets (N=1681). Here we report the statistical significance of ReCreate's improvements over the best baseline in each of the three baseline categories, including human-designed, self-evolve, and agent generation.
\begin{table}[t]
\centering
\caption{Paired significance tests (ReCreate vs baselines).}
\label{tab:r3}
\begin{tabular}{lcc}
\toprule
\textbf{Comparison} & \textbf{exact $p$} & \textbf{significant?} \\
\midrule
 vs SBA & 3.43e-11 & \checkmark \\
 vs ExpeL & 7.57e-8 & \checkmark \\
 vs AgentSquare & 9.24e-11 & \checkmark \\
\bottomrule
\end{tabular}
\end{table}

The results show that ReCreate achieves statistically significant ($\alpha = 0.05$) gains over these baselines.

\appsection{Sensitivity Analysis}

To address the concern about dev-set sensitivity, we conduct two sensitivity analyses as follows.

\paragraph{Sensitivity of Development Set} We construct dev sets by randomly sampling 20/50/100 cases from SWE-bench Full excluding SWE-bench Verified (500 cases), and keep all other settings identical to the main experiment. The results are as follows:

\begin{table}[t]
\centering
\caption{Changing the size of dev set in SWE tasks.}
\label{tab:r4}
\begin{tabular}{lc}
\toprule
\textbf{Method} & \textbf{SWE-bench Verified} \\
\midrule
ReCreate-20 & 57.4 \\
ReCreate-50 & 60.4 \\
ReCreate-100 & 60.8 \\
\bottomrule
\end{tabular}
\end{table}

We conclude that the performance of ReCreate improves as the dev set size increases. This is consistent with our expectations since a larger dev set provides more experience for agent scaffold updates. In our experiments, we use only a small development set of 20 samples and can beat various compared baselines. In fact, ReCreate still has substantial room for improvement and strong potential when larger dev sets are used.

\paragraph{Sensitivity of Dev Set Composition} We construct dev sets by repeatedly and randomly sampling 50 instances, thereby creating three dev sets with different compositions. The instances of dev sets are sampled from SWE-bench Full excluding SWE-bench Verified and all other settings are identical to the main experiment. The results of ReCreate are as follows:

\begin{table}[t]
\centering
\caption{Changing the composition of dev set in SWE tasks.}
\label{tab:r5}
\begin{tabular}{lc}
\toprule
\textbf{Method} & \textbf{SWE-bench Verified} \\
\midrule
ReCreate-dev1 & 60.4 \\
ReCreate-dev2 & 59.8 \\
ReCreate-dev3 & 59.8 \\
\bottomrule
\end{tabular}
\end{table}

Even with different dev-set samples, performance remains stable (std = 0.28), suggesting that \textbf{ReCreate is not sensitive to dev-set composition.}

In our main experiments, we use a dev set of 20 instances as a compute-efficient choice, and this dev-set size is also common in prior works, such as~\citep{hu2024automated, shang2024agentsquare}.

\appsection{Cases in Motivation}\label{Case Studies in Motivation}

In this section we present cases that show experience can be important for agent scaffold update, shown in Figure~\ref{fig:case_rules},~\ref{fig:case_tools},~\ref{fig:case_workflows}.

\begin{figure*}[t]
    \centering
    \begin{codepromptbox}{Motivation 1: Rules (Log: ML/q\_39)}
\small
\textbf{Context:} Task requires training a classifier on \texttt{breast\_cancer.csv}.

\vspace{0.2em}
\textbf{[STEP 3] Model Training \& Verification}

\vspace{0.2em}
\textbf{> THOUGHT:}
...missing values are handled. Features are stored in \texttt{X} and labels in \texttt{y}. I will now fit a Random Forest and check the accuracy to ensure the model is learning correctly before prediction.

\vspace{0.2em}
\textbf{> ACTION:}
\begin{verbatim}

from sklearn.ensemble import RandomForestClassifier
...
# Initialize model with default parameters
clf = RandomForestClassifier(random_state=42)
clf.fit(X, y)
...
# Evaluate performance
# <--- [FATAL FLAW] Evaluating on training data
train_score = clf.score(X, y)
print(f"Current Accuracy: {train_score:.4f}")
\end{verbatim}

\textbf{> OBSERVATION:}
\texttt{Current Accuracy: 0.9420}

\vspace{0.2em}
\textbf{> THOUGHT:}
The accuracy is 94.2\%, which is excellent. The model is robust. I will proceed to load \texttt{test.csv}...

\vspace{0.5em}
\hrule
\vspace{0.5em}

\textbf{[Motivation]}
The trace reveals an overfitting hallucination.
The agent sees a near-perfect score (0.942) and stops optimizing, unaware that the score is biased without a hold-out set.
A naive idea is to inject rule: ``CONSTRAINT: You need to use \texttt{train\_test\_split} for internal evaluation.''
    \end{codepromptbox}
    \caption{A case study for adding rules.}
    \label{fig:case_rules}
\end{figure*}

\begin{figure*}[htbp]
    \centering
    \begin{codepromptbox}{Motivation 2: Tools (Log: SA/q\_41)}
\small
\textbf{Context:} Agent generated \texttt{solution.py} and needs to verify it.

\vspace{0.2em}
\textbf{[STEP 4] Multi-step Verification}

\vspace{0.2em}
\textbf{> THOUGHT:}
...code is written. Before submission, I must ensure the file is not empty, has correct syntax, and is actually executable. I will check these one by one.

\vspace{0.2em}
\textbf{> ACTION:}
...
\begin{verbatim}
# 1. Check if file exists and is not empty
ls -lh solution.py && cat solution.py | head -n 3
\end{verbatim}

\textbf{> OBSERVATION:}
\texttt{-rw-r--r-- 1 root root 402B ...} \\
\texttt{def solve(nums): ...}

\vspace{0.2em}
\textbf{> ACTION:}
\begin{verbatim}
# 2. Check for syntax errors
python3 -m py_compile solution.py
\end{verbatim}

\textbf{> OBSERVATION:}
(No output, implies syntax OK)

\vspace{0.2em}
\textbf{> ACTION:}
\begin{verbatim}
# 3. Check execution logic
# <--- [INEFFICIENT] Wasting 3 turns on routine checks
python3 -c "from solution import solve; print(solve([1, 2]))"
\end{verbatim}

\textbf{> OBSERVATION:}
\texttt{3}

\vspace{0.5em}
\hrule
\vspace{0.5em}

\textbf{[Motivation]}
The trace reveals a tedious verification loop. The agent spends multiple turns performing generic sanity checks (size, syntax, execution).
An effective idea is to create a unified tool: \texttt{full\_check(file\_path)} that automates this entire validation pipeline in one call.
    \end{codepromptbox}
    \caption{A case study for adding tools.}
    \label{fig:case_tools}
\end{figure*}

\begin{figure*}[t]
    \centering
    \begin{codepromptbox}{Motivation 3: Workflows (Log: SWE-bench/django-45)}
\small
\textbf{Context:} Fixed bug in \texttt{django/utils/text.py}. Tests passed.

\vspace{0.2em}
\textbf{[STEP 12] Submission}

\vspace{0.2em}
\textbf{> THOUGHT:}
...regression tests are passing. I will now commit the changes to the repository and submit the final patch.

\vspace{0.2em}
\textbf{> ACTION:}
\begin{verbatim}
git add django/utils/text.py
# <--- [FATAL FLAW] Committing clears the Index
git commit -m "Fix regression in slugify"
\end{verbatim}

\textbf{> OBSERVATION:}
\begin{verbatim}
[main 8a3c1b] Fix regression in slugify
 1 file changed, 4 insertions(+), 1 deletion(-)
\end{verbatim}

\vspace{0.2em}
\textbf{> ACTION:}
\begin{verbatim}
echo COMPLETE_TASK_AND_SUBMIT_FINAL_OUTPUT
\end{verbatim}

\textbf{> SYSTEM EVALUATION:}
\texttt{>> Patch Content:} (empty string) \\
\texttt{>> Result: FAILED (Patch size 0 bytes)}

\vspace{0.5em}
\hrule
\vspace{0.5em}

\textbf{[Motivation]}
The trace reveals a ``clean working tree'' paradox.
The agent followed standard Git practice, but the harness requires staged changes (Index) for patch extraction.
A necessary update is to enforce a workflow: ``CRITICAL: RUN \texttt{git diff --cached} before finish.''
    \end{codepromptbox}
    \caption{A case study for enforcing workflows.}
    \label{fig:case_workflows}
\end{figure*}

\appsection{Prompts of ReCreate-Agent}
\label{app:system_prompt}

The ReCreate-Agent operates as an agent-optimizer.
Its system prompt is designed to guide it through the full loop of inspection, diagnosis, and scaffold evolution. Below we present the core components of the prompt (administrative instructions and specific file paths are omitted for brevity), shown in Figure~\ref{fig:system_prompt}.

\begin{figure*}[t]
    \centering
\begin{codepromptbox}{ReCreate-Agent System Prompt}
\small
\textbf{Role Definition}
You are an agent that creates and evolves other AI agents by editing their scaffolds (prompts, workflows, tools, and memory mechanisms).

\textbf{Mission}
Analyze agent execution trajectories, understand success and failure patterns, inspect the agent's environment, and evolve the agent's scaffold and tools so that it performs better on future tasks in the same domain.

\hrulefill

\textbf{Core Philosophy}
You are discovering \textbf{generalizable principles}. Think like a teacher improving a student:
\begin{itemize}
    \item \textbf{Learn from SUCCESS}: Extract winning strategies and encode them as tools.
    \item \textbf{Learn from FAILURE}: Diagnose issues and add safeguards.
\end{itemize}

\hrulefill

\textbf{The Five Components You Control}
\begin{enumerate}
    \item \texttt{system\_template}: Agent's identity, core knowledge, principles.
    \item \texttt{instance\_template}: Problem-solving workflow, step-by-step guidance.
    \item \texttt{memory\_template}: Agent's memory read/write strategy.
    \item \texttt{agent\_tools/}: Reusable automation scripts and helper commands.
    \item \texttt{agent\_memory/}: Historical lessons \& patterns (static content).
\end{enumerate}

\hrulefill

\textbf{Thinking Framework}
When analyzing a trajectory, focus on:
\begin{itemize}
    \item \textbf{Patterns}: What behaviors systematically help or hinder progress?
    \item \textbf{Root Cause}: Is this a knowledge gap, strategy gap, or tool gap?
    \item \textbf{Intervention}: What targeted change would steer future trajectories better?
    \item \textbf{Tool Opportunities}: Ask "What repetitive operation could be automated?"
\end{itemize}

\hrulefill

\textbf{Available Tools (Action Space)}
\begin{itemize}
    \item \textbf{Trajectory Analysis}:
    \begin{itemize}
        \item \texttt{read\_trajectory.py summary}: Get overview of the run.
        \item \texttt{read\_trajectory.py failures}: List all errors and their context.
        \item \texttt{read\_trajectory.py context --step N}: Inspect specific reasoning steps.
        \item ......
    \end{itemize}
    
    \item \textbf{Environment Inspection}:
    \begin{itemize}
        \item \texttt{inspect\_in\_docker.py --command "ls -R"}: View the actual file system state.
        \item ......
    \end{itemize}
    
    \item \textbf{Scaffold Editing}:
    \begin{itemize}
        \item \texttt{scaffold\_editor.py str\_replace}: Modify prompts/templates.
        \item \texttt{tool\_manager.sh create}: Create new Python tools for the agent.
        \item \texttt{memory\_manager.py add}: Inject static knowledge/lessons.
        \item ......
    \end{itemize}
\end{itemize}

\hrulefill

\textbf{Recommended Workflow}
\begin{itemize}
    \item \textbf{Check Submission}: Verify if the patch is empty or valid.
\item \textbf{Read Trajectory}: Understand what the agent did step by step.
\item \textbf{Analyze Causes}: Why did the reasoning or tool usage break down?
\item \textbf{Decide Intervention}: Create a tool (Preferred) or Update Scaffold.
\item \textbf{Execute \& Verify}: Apply changes and confirm they match intent.
\end{itemize}
\end{codepromptbox}
    \caption{Main prompt for ReCreate-Agent.}
    \label{fig:system_prompt}
\end{figure*}

The Synthesis prompt of Meta-Agent aggregates per-instance scaffold edits into one unified scaffold, shown in Figure~\ref{fig:batch_synthesis_prompt}.
We omit administrative instructions and absolute paths for brevity.

\begin{figure*}[t]
    \centering
\begin{codepromptbox}{Synthesis Meta-Agent Prompt (Batch Synthesis)}
\small

\textbf{Role}\\
You are the \textbf{Synthesis Meta-Agent}. Review all proposed scaffold modifications from a batch and produce a unified \emph{global} scaffold that generalizes across the domain.

\textbf{Context}\\
\textbf{Batch \texttt{\{\{ batch\_idx \}\}}} ran \texttt{\{\{ num\_instances \}\}} instances from the same base scaffold (\texttt{global\_v\{\{ batch\_idx \}\}}). \texttt{\{\{ num\_modifications \}\}} instances proposed modifications.

\hrulefill

\textbf{Your Task}\\
\begin{enumerate}
    \item Review each proposal (summary + diff; open full files when needed).
    \item Extract shared patterns and generalizable improvements.
    \item Resolve conflicts and synthesize a single unified scaffold.
\end{enumerate}

\hrulefill

\textbf{Where to Inspect Full Proposals}\\
\begin{itemize}
    \item \texttt{batch\_modifications/<instance\_id>/diff.txt}
    \item \texttt{batch\_modifications/<instance\_id>/summary.md}
    \item \texttt{batch\_modifications/<instance\_id>/scaffold.yaml}
\end{itemize}

\hrulefill

\textbf{Decision Guidelines (Prefer Success)}\\
\begin{itemize}
    \item \textbf{Successful instances:} prioritize reusable tools, stable workflow improvements, and concise rules that clearly contributed to success.
    \item \textbf{Failed instances:} include only low-risk safeguards that address a \emph{general} failure mode; avoid brittle or overly restrictive rules.
\end{itemize}

\textbf{Conflict Resolution}\\
Prefer changes supported by multiple instances; otherwise choose the simpler, more general formulation. When uncertain, keep the original rule.

\hrulefill

\textbf{Required Outputs}\\
\begin{itemize}
    \item Update \texttt{current/scaffold.yaml} with the unified scaffold.
    \item Write \texttt{current/synthesis\_summary.md} (included vs. rejected changes, plus conflict resolutions).
    \item If useful tools/memories exist, copy/add them into \texttt{current/}.
\end{itemize}

\textbf{Completion}\\
Output: \texttt{echo COMPLETE\_TASK\_AND\_SUBMIT\_FINAL\_OUTPUT}

\hrulefill

\textbf{Reminder}\\
Optimize for \textbf{generalization across the domain}, not for the specific instances in this batch.

\end{codepromptbox}
    \caption{Batch synthesis prompt for aggregating instance-level scaffold edits into a unified global scaffold.}
    \label{fig:batch_synthesis_prompt}
\end{figure*}

\appsection{Case Study}\label{Case Study}
We take the initialization and the final results of ReCreate on Django for demonstration.
To ensure a realistic initialization, ReCreate starts with a minimal seed scaffold, shown in Figure~\ref{fig:seed_scaffold}. Driven by interaction experience, the agent iteratively evolves this seed into a specialized domain system. The final output includes rigorous prompt templates for system constraints, workflows, and memory interfaces (Figures~\ref{fig:prompt_system}, \ref{fig:prompt_instance}, \ref{fig:prompt_memory}). Additionally, ReCreate crystallizes its experience into actionable memories (Figure~\ref{fig:generated_memories}) and custom tools (Figure~\ref{fig:generated_tool}) to resolve specific domain challenges.

\begin{figure*}[t]
    \centering

    \begin{tcolorbox}[promptstyle, title={[System Template] Minimal Seed}]
    You are an expert software engineer solving GitHub issues.
    
    \textbf{Response Format:}
    THOUGHT: <analysis>
    \vspace{-0.5em}
    \begin{verbatim}
```bash
<ONE command>

\end{verbatim}
\vspace{-0.8em}

\textbf{Rules:}
\begin{itemize}[leftmargin=*, nosep]
    \item ONE command per response. Do NOT try to do everything at once.
    \item No vim/nano. Use \texttt{sed -i} or \texttt{python3 -c} for edits.
    \item NEVER use heredoc (<<EOF) - it causes truncation errors.
    \item NEVER read or modify test files.
\end{itemize}
\end{tcolorbox}

\vspace{0.2em}

\begin{tcolorbox}[promptstyle, title={[Instance Template] Minimal Seed}]
\textbf{\#\# Task} \\
\{\{task\}\}

\textbf{\#\# Workflow (STEP BY STEP)}
\begin{enumerate}[leftmargin=*, label={\arabic*.}, nosep]
    \item LOCATE: Find relevant files with \texttt{find} and \texttt{grep}.
    \item ANALYZE: Read the code...
    \item IMPLEMENT: Edit the files...
    \item VERIFY: Check if it works...
    \item SUBMIT: \texttt{git add -A \&\& git diff --cached \&\& echo COMPLETE\_TASK...}
\end{enumerate}
\end{tcolorbox}

\caption{The Minimal Seed Scaffold in Django.}
\label{fig:seed_scaffold}
\end{figure*}

\begin{figure*}[h]
    \centering
    \begin{tcolorbox}[
        colback=gray!5, colframe=gray!60, boxrule=0.8pt, arc=2mm,
        title=\textbf{Created System Template},
        fonttitle=\bfseries\sffamily, fontupper=\scriptsize\ttfamily,
        left=3mm, right=3mm, top=3mm, bottom=3mm
    ]
    You are an expert software engineer solving GitHub issues in real open-source projects.

    \vspace{0.5em}
    \textbf{\#\# Response Format (CRITICAL)}\\
    You MUST respond with EXACTLY this format every turn:

    THOUGHT: <your analysis in a single paragraph>

    ```bash\\
    <exactly ONE command>\\
    ```

    NEVER output multiple code blocks. NEVER skip the THOUGHT section.

    \vspace{0.5em}
    \textbf{\#\# Rules}
    \begin{itemize}
        \setlength\itemsep{0em}
        \item ONE command per response. Work step-by-step, not all at once.
        \item No interactive tools: NO vim, nano, less, or interactive prompts
        \item NEVER use heredoc (<<EOF) - it causes truncation errors. Use sed or python -c instead.
        \item Directory changes don't persist. Use: cd /path \&\& command
        \item Commands timeout after 100 seconds. Use head/tail/grep for large outputs.
    \end{itemize}

    \vspace{0.5em}
    \textbf{\#\# Anti-Cheat Rules (MUST FOLLOW)}
    \begin{itemize}
        \setlength\itemsep{0em}
        \item NEVER read, modify, or look at any test files (test\_*.py, *\_test.py, tests/, testing/)
        \item NEVER try to discover or reverse-engineer test content
        \item NEVER modify or delete .git directories
        \item Focus ONLY on the issue description to understand what needs fixing
        \item Your fix must be based on understanding the problem, not the tests
    \end{itemize}

    \vspace{0.5em}
    \textbf{\#\# File Editing Methods}\\
    RECOMMENDED: Use the file\_editor tool (handles quoting automatically):\\
    \texttt{python3 /workspace/agent\_tools/utils/file\_editor/main.py replace <file> --old "old text" --new "new text"}\\
    \texttt{python3 /workspace/agent\_tools/utils/file\_editor/main.py show <file> --start 50 --end 100}\\
    \texttt{python3 /workspace/agent\_tools/utils/file\_editor/main.py search <file> --pattern "regex"}

    For simple changes, use sed:\\
    \texttt{sed -i 's/old\_text/new\_text/g' filename.py}

    For complex edits with special characters, use python with raw strings:\\
    \texttt{python3 -c 'from pathlib import Path; p=Path("file.py"); s=p.read\_text(); s=s.replace("old", "new"); p.write\_text(s)'}

    \vspace{0.5em}
    \textbf{\#\# Running Python Scripts}\\
    For complex Python scripts (multi-line, classes, if statements), use printf to write to a file:\\
    \texttt{printf 'line1\textbackslash nline2\textbackslash nprint(result)' > /tmp/test.py \&\& python3 /tmp/test.py}

    \vspace{0.5em}
    \textbf{\#\# Codebase Location}\\
    The repository is at /testbed/. Always work from there.
    \end{tcolorbox}
    \caption{The System Template created from Django experience.}
    \label{fig:prompt_system}
\end{figure*}

\begin{figure*}[t]
    \centering
    \begin{tcolorbox}[
        colback=gray!5, colframe=gray!60, boxrule=0.8pt, arc=2mm,
        title=\textbf{Created Instance Template},
        fonttitle=\bfseries\sffamily, fontupper=\scriptsize\ttfamily,
        left=3mm, right=3mm, top=3mm, bottom=3mm
    ]
    \textbf{\#\# GitHub Issue to Solve}\\
    \{\{task\}\}

    \vspace{0.5em}
    \textbf{\#\# Workflow (Follow Step-by-Step)}
    
    1. UNDERSTAND: Read the issue carefully. What is the expected vs actual behavior?
    
    2. LOCATE: Find relevant files using:
    \begin{itemize}[leftmargin=1.5em]
        \item \texttt{find /testbed -type f -name "*.py" | grep -E "keyword" | head -20}
        \item \texttt{grep -r "function\_name" /testbed --include="*.py" -l | head -20}
    \end{itemize}
    If grep output is truncated, NARROW your search to the relevant subdirectory:
    \begin{itemize}[leftmargin=1.5em]
        \item \texttt{grep -r "pattern" /testbed/specific/module/ --include="*.py"}
    \end{itemize}
    After finding one occurrence, check if the same pattern exists in related files.
    
    3. ANALYZE: Read and understand the code:
    \begin{itemize}[leftmargin=1.5em]
        \item \texttt{cat /testbed/path/to/file.py | head -100}
        \item \texttt{grep -n "pattern" /testbed/path/to/file.py}
    \end{itemize}
    
    4. IMPLEMENT: Make targeted changes using sed or python -c (NO heredoc!)
    
    5. VERIFY: Check your changes:
    \begin{itemize}[leftmargin=1.5em]
        \item \texttt{git diff} to see what you changed
        \item Quick sanity check: \texttt{python3 -c "import module\_you\_changed"} to verify no syntax errors
        \item TEST THE ACTUAL BEHAVIOR: Run a quick test with the specific inputs from the issue\\
          Example: \texttt{python3 -c "from module import func; print(func(problematic\_input))"}
        \item Test edge cases: boundary values, empty inputs, unusual but valid inputs
        \item For multi-stage operations (ORM queries, serialization, caching): trace the full data flow and verify your fix handles ALL stages, not just the first one you found
    \end{itemize}
    
    6. VALIDATE: Before submitting, ask yourself:
    \begin{itemize}[leftmargin=1.5em]
        \item Does my fix address the root cause described in the issue
        \item Did I test the specific scenario mentioned in the issue?
        \item Could my change break other functionality?
    \end{itemize}
    
    7. SUBMIT: When confident your fix is complete:\\
    \texttt{git add -A \&\& git diff --cached \&\& echo COMPLETE\_TASK\_AND\_SUBMIT\_FINAL\_OUTPUT}

    \vspace{0.5em}
    \textbf{\#\# Important Reminders}
    \begin{itemize}
        \setlength\itemsep{0em}
        \item Work incrementally: one change at a time
        \item Read code before modifying it
        \item Verify your changes with git diff before submitting
        \item NEVER touch test files - your fix will be evaluated against hidden tests
    \end{itemize}
    \end{tcolorbox}
    \caption{The Instance Template created from Django experience.}
    \label{fig:prompt_instance}
\end{figure*}

\begin{figure*}[h]
    \centering
    \begin{tcolorbox}[
        colback=gray!5, colframe=gray!60, boxrule=0.8pt, arc=2mm,
        title=\textbf{Created Memory Template},
        fonttitle=\bfseries\sffamily, fontupper=\scriptsize\ttfamily,
        left=3mm, right=3mm, top=3mm, bottom=3mm
    ]
    \textbf{\#\# Memory System}\\
    You can read and write memories to learn from past experiences.
    
    Read memories (search for relevant lessons):\\
    \texttt{python3 /workspace/agent\_memory/search\_memory.py "keyword"}
    
    Write memories (save what you learned):\\
    \texttt{python3 /workspace/agent\_memory/write\_memory.py --title "Short title" --content "What you learned" --tags "tag1,tag2"}
    
    \vspace{0.5em}
    When to read memories:
    \begin{itemize}
        \setlength\itemsep{0em}
        \item At the start of a task, search for similar issue types
        \item When encountering an error, search for that error message
        \item When stuck, search for the technology/library involved
    \end{itemize}
    
    When to write memories:
    \begin{itemize}
        \setlength\itemsep{0em}
        \item After solving a tricky bug, save the key insight
        \item When you discover a useful pattern or gotcha
        \item After finding a non-obvious solution approach
    \end{itemize}
    \end{tcolorbox}
    \caption{The Memory Template created from Django experience.}
    \label{fig:prompt_memory}
\end{figure*}

\begin{figure*}[t]
    \centering
    \begin{tcblisting}{
        colback=gray!5, 
        colframe=gray!60, 
        boxrule=0.8pt, 
        arc=2mm,
        title=\textbf{[Generated Memories] agent\_memory.yaml},
        fonttitle=\bfseries\sffamily,
        listing only,
        listing options={style=yamlstyle}, % 使用 YAML 样式
        left=1mm, right=1mm, top=1mm, bottom=1mm,
        enhanced
    }
memories:
- content: "When using python3 -c for file editing with regex patterns, avoid complex escaping. Instead use raw strings (r'pattern') and multi-line format. Better yet, use the file_editor tool at /workspace/agent_tools/utils/file_editor/main.py for safe search-and-replace operations."
  created: '2026-01-02'
  id: mem_001
  tags: [python, regex, editing, escaping]
  title: Regex escaping in python -c

- content: "For Django issues, always run the specific FAIL_TO_PASS tests mentioned in expected_tests.txt before submitting. Use: python tests/runtests.py <module>.<Class>.<method>. Don't assume a fix works just because it compiles - the actual test may reveal the fix is incomplete."
  created: '2026-01-02'
  id: mem_002
  tags: [django, testing, swe-bench, verification]
  title: Run FAIL_TO_PASS tests before submitting Django fixes

- content: "When replacing entire Python methods/functions, avoid complex regex patterns that require escaping. Instead, use str.find() to locate the method start (e.g., 'def method_name(') and find the next method at the same indentation level to determine the end. This approach is more reliable."
  created: '2026-01-02'
  id: mem_003
  tags: [python, editing, methods, refactoring]
  title: "Method replacement: use find() not regex"
    \end{tcblisting}
    \caption{A snapshot of the static memory accumulated by ReCreate.}
    \label{fig:generated_memories}
\end{figure*}

\begin{figure*}[t]
    \centering
    \begin{tcblisting}{
        colback=gray!5, 
        colframe=gray!60, 
        boxrule=0.8pt, 
        arc=2mm,
        title=\textbf{[Created Tool] replace\_method.py},
        fonttitle=\bfseries\sffamily,
        listing only,             % 告诉 tcolorbox 里面只放代码
        listing options={style=recreatepython}, % 使用上面定义的样式
        left=1mm, right=1mm, top=1mm, bottom=1mm,
        enhanced
    }
#!/usr/bin/env python3
"""
Replace Method Tool - Replace an entire Python method/function by name

Usage:
    python3 replace_method/main.py <file> --method "name" --new-body "def name(): pass"
    python3 replace_method/main.py <file> --method "name" --new-body-file /path/to/impl.py
"""

import argparse, re, sys
from pathlib import Path

def find_method_boundaries(content: str, method_name: str) -> tuple:
    """Find the start and end positions of a method based on indentation."""
    lines = content.split('\n')
    
    # Pattern to match method/function definition strictly
    method_pattern = re.compile(rf'^(\s*)(def\s+{re.escape(method_name)}\s*\()')
    
    # ... [Logic to find start_line and base_indent] ...
    
    # Find the end of the method by checking indentation levels
    end_line = len(lines)
    for i in range(start_line + 1, len(lines)):
        line = lines[i]
        stripped = line.lstrip()
        
        # Skip empty lines and comments
        if not stripped or stripped.startswith('#'): continue
        
        current_indent = len(line) - len(stripped)
        
        # Stop if we find a line at same or lower indentation level
        if current_indent <= base_indent:
            end_line = i
            break
            
    return start_pos, end_pos, base_indent

def replace_method(filepath: str, method_name: str, new_body: str) -> None:
    """Replace a method/function with new implementation."""
    path = Path(filepath)
    content = path.read_text()
    
    # 1. Locate the method in the source file
    start_pos, end_pos, indent = find_method_boundaries(content, method_name)
    
    if start_pos is None:
        print(f"ERROR: Method '{method_name}' not found")
        sys.exit(1)
    
    # 2. Align the new body to the correct indentation level
    new_lines = new_body.split('\n')
    indented_lines = []
    for line in new_lines:
        if line.strip() and not line.startswith(' ' * indent):
            line = ' ' * indent + line.lstrip()
        indented_lines.append(line)
    new_body_indented = '\n'.join(indented_lines)
    
    # 3. Perform the string replacement
    new_content = content[:start_pos] + new_body_indented + content[end_pos:]
    path.write_text(new_content)
    print(f"SUCCESS: Replaced method '{method_name}' in {filepath}")

def main():
    # ... [Argparse setup omitted for brevity] ...
    if args.show_only:
        replace_method(args.file, args.method, "", show_only=True)
    elif args.new_body:
        replace_method(args.file, args.method, args.new_body)

if __name__ == "__main__":
    main()
    \end{tcblisting}
    \caption{A tool created from Django experience.}
    \label{fig:generated_tool}
\end{figure*}

\end{document}